%% file: main.tex
\documentclass[compsoc]{IEEEtran}
\usepackage{amsmath,amsfonts}
\usepackage{array}
\usepackage[caption=false,font=normalsize,labelfont=sf,textfont=sf]{subfig}
\usepackage{textcomp}
\usepackage{stfloats}
\usepackage{url}
\usepackage{verbatim}
\usepackage{graphicx}
\def\BibTeX{{\rm B\kern-.05em{\sc i\kern-.025em b}\kern-.08em
    T\kern-.1667em\lower.7ex\hbox{E}\kern-.125emX}}
\usepackage{balance}

\usepackage{tabu}                      % only used for the table example
\usepackage{booktabs}                  % only used for the table example
\usepackage{lipsum}                    % used to generate placeholder text
\usepackage{mwe}                       % used to generate placeholder figures
\usepackage{amsmath}
\usepackage{makecell}

\usepackage{mathptmx}                  % use matching math font
\usepackage{hyperref}

\usepackage{algpseudocode}
\usepackage{algorithm}
\usepackage[dvipsnames]{xcolor}
\usepackage{colortbl}

\usepackage{float}

\usepackage{soul}

\newcommand{\rev}[1]{#1}

\newcommand{\boldsubsubsection}[1]{
\vspace{4pt}
\noindent
\textbf{#1.}
}

\begin{document}

\title{UMATO: Bridging Local and Global Structures for Reliable Visual Analytics with \\Dimensionality Reduction}
\author{Hyeon Jeon, Kwon Ko, Soohyun Lee, Jake Hyun, Taehyun Yang, Gyehun Go, Jaemin Jo, and Jinwook Seo
\IEEEcompsocitemizethanks{
    \IEEEcompsocthanksitem Hyeon Jeon, Soohyun Lee, Jake Hyun, Taehyun Yang, Gyehun Go, and Jinwook Seo are with Seoul National University.
  	E-mail: \{hj, shlee\}@hcil.snu.ac.kr, 
   \{jakehyun, 0705danny, rotation\_430, jseo\}@snu.ac.kr
    \IEEEcompsocthanksitem Kwon Ko is with Stanford University. E-mail: kwonko@stanford.edu
    \IEEEcompsocthanksitem Jaemin Jo is with Sungkyunkwan University. E-mail: jmjo@skku.edu
    \IEEEcompsocthanksitem Jaemin Jo and Jinwook Seo are corresponding authors.
    \protect\\
}}

\markboth{IEEE Transactions on Visualization and Computer Graphics}%
{Paper short title}

\IEEEtitleabstractindextext{
\begin{abstract}
\input{sections/00_abstract}

\end{abstract}
}

\maketitle
\IEEEdisplaynontitleabstractindextext

\input{sections/01_introduction}

\input{sections/02_related_works}

\input{sections/03_umato}

\input{sections/04_implementation}

\input{sections/05_experiments}

\input{sections/06_demonstration}

\input{sections/07_use_case}

\input{sections/08_hyperparameter}

\input{sections/09_discussions}
\input{sections/10_conclusion}

\section*{Acknowledgments}
This work was supported by the National Research Foundation of Korea (NRF) grant funded by the Korea government (MSIT) (No. RS-2023-NR077081 and No. RS-2023-00221186), and the SNU-Global Excellence Research Center establishment project.
This work was also supported by the
Institute of Information \& communications Technology Planning
\& Evaluation (IITP) grant funded by the Korea government (MSIT)
[NO.2021-0-01343, Artificial Intelligence Graduate School Program
(Seoul National University)]. 
The ICT at
Seoul National University provided research facilities for this study.
Hyeon Jeon is in part supported by Google Ph.D. Fellowship. 

% This work was supported by the National Research Foundation of Korea (NRF) grant funded by the Korea government(MSIT)(No. RS-2023-00221186).

\bibliographystyle{IEEEtran}
\bibliography{ref}

\newcommand{\spacingv}{\vspace{-3mm}}

\begin{IEEEbiography}[{\includegraphics [width=1in,height=1.25in,clip, keepaspectratio]{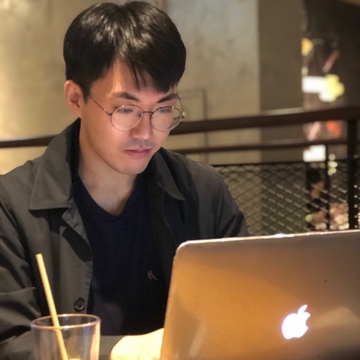}}] {Hyeon Jeon} 
is a Ph.D. Student at the Department of Computer Science and Engineering, Seoul National University. His research interests span the field of Visual Analytics and Machine Learning. Before starting his Ph.D. program, he received a B.S. degree in Computer Science and Engineering from POSTECH. 
\end{IEEEbiography}

\spacingv

\begin{IEEEbiography}[{\includegraphics [width=1in,height=1.25in,clip, keepaspectratio]{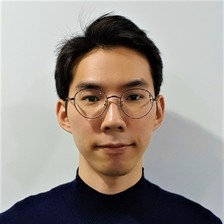}}] {Kwon Ko} 
is a Ph.D. Student at Stanford University. Prior, he received a B.S. degree in Mathematics from Hanyang University and received an M.S. degree in Computer Science and Engineering from Seoul National University.
\end{IEEEbiography}

\spacingv

\begin{IEEEbiography}[{\includegraphics [width=1in,height=1.25in,clip, keepaspectratio]{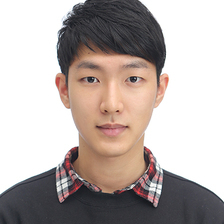}}] {Soohyun Lee} 
is a Ph.D. Student at the Department of Computer Science and Engineering, Seoul National University. Before starting his Ph.D. program, he received a B.S. degree in Computer Science and Engineering from the Korea University, Seoul, Korea. 
\end{IEEEbiography}

\spacingv

\begin{IEEEbiography}[{\includegraphics [width=1in,height=1.25in,clip, keepaspectratio]{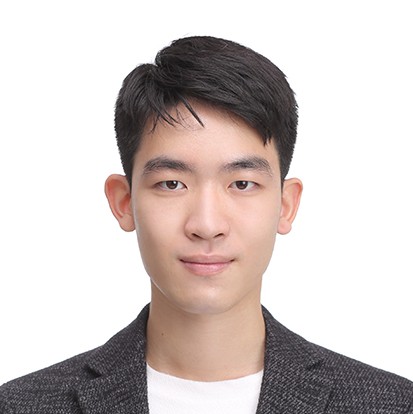}}] {Jake Hyun} is an incoming Ph.D. student in Computer Science at Cornell University. He received his B.S. in Computer Science and Engineering with a minor in Linguistics from Seoul National University. His research focuses on designing efficient computing systems, with an emphasis on hardware-software co-design and scalable machine learning.
\end{IEEEbiography}

\spacingv

\begin{IEEEbiography}[{\includegraphics [width=1in,height=1.25in,clip, keepaspectratio]{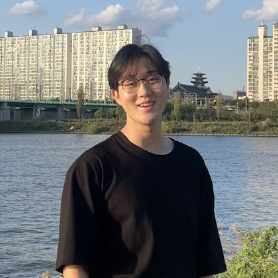}}] {Taehyun Yang}
is a Ph.D. Student at the Department of Computer Science, University of Maryland. Before starting his Ph.D. program, he received a B.S. degree in Computer Science and Engineering from Seoul National University, Seoul, Korea.
\end{IEEEbiography}

\spacingv

\begin{IEEEbiography}[{\includegraphics [width=1in,height=1.25in,clip, keepaspectratio]{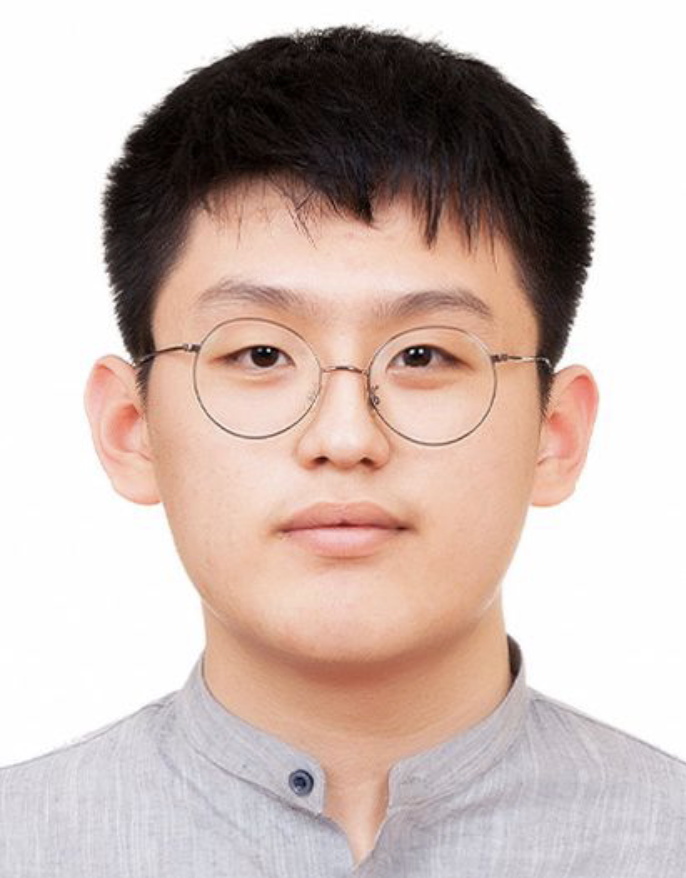}}] {Gyehun Go}
is an undergraduate student at the Department of Computer Science and Engineering at Seoul National University, Seoul.
\end{IEEEbiography}

\spacingv

\begin{IEEEbiography}[{\includegraphics [width=1in,height=1.25in,clip, keepaspectratio]{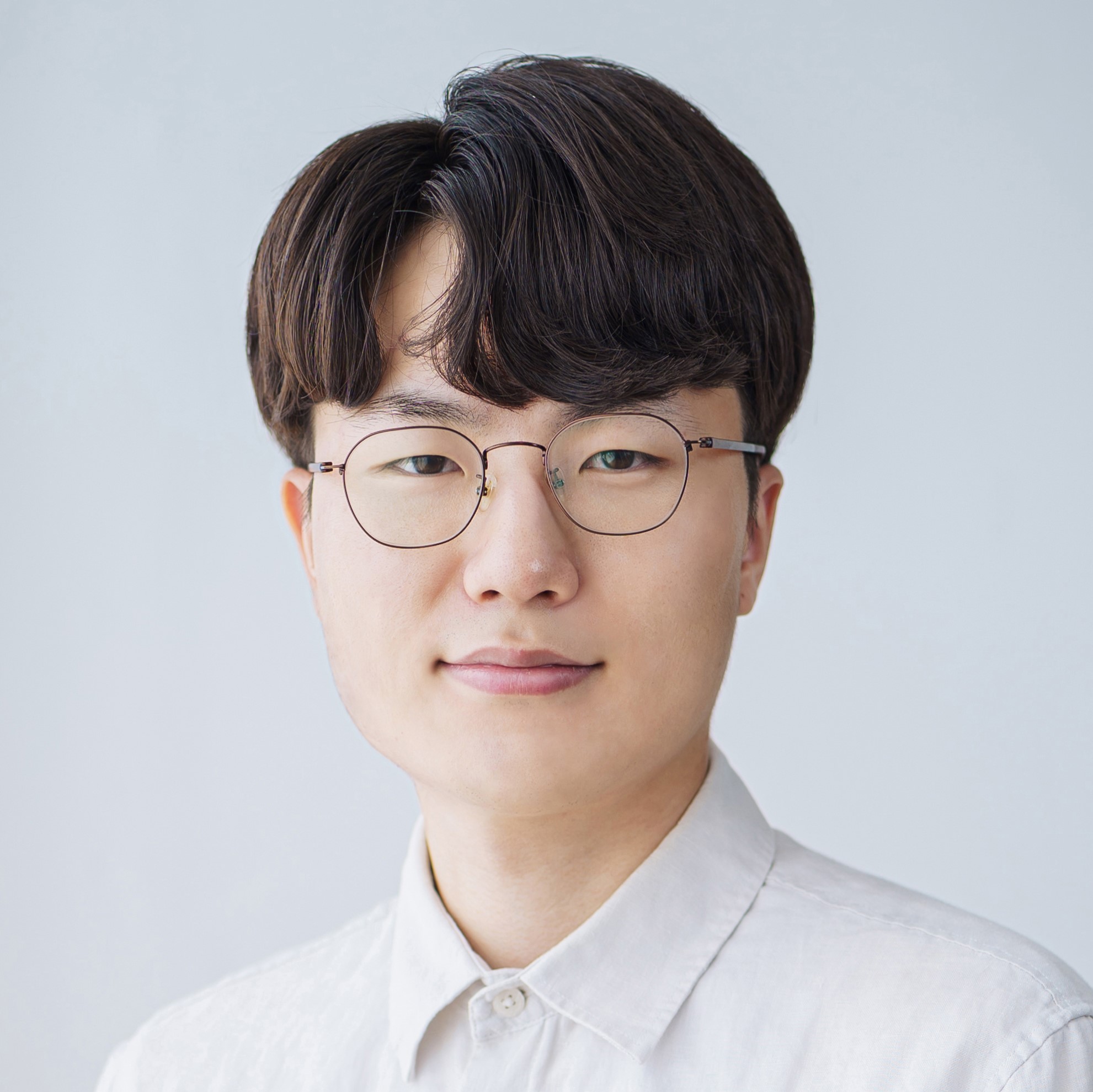}}] {Jaemin Jo} 
received the BS and PhD degrees in computer science and engineering from Seoul National University, Seoul, South Korea, in 2014 and 2020, respectively. He is currently an associate professor with the College of Computing and Informatics, Sungkyunkwan University, Korea. His research interests include human-computer interaction
and large-scale data visualization.
\end{IEEEbiography}

\spacingv

\begin{IEEEbiography}[{\includegraphics [width=1in,height=1.25in,clip, keepaspectratio]{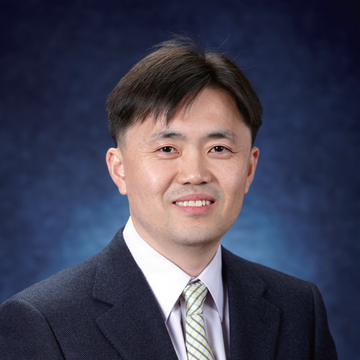}}] {Jinwook Seo} 
is a professor in the Department of Computer Science and Engineering, Seoul National University, where he is also the Director of the Human-Computer Interaction Laboratory. His research interests include Human-Computer Interaction, Information Visualization, and Biomedical Informatics. He received his PhD in Computer Science from the University of Maryland at College Park in 2005.
\end{IEEEbiography}

\end{document}

%% file: sections/00_abstract.tex
Due to the intrinsic complexity of high-dimensional (HD) data, dimensionality reduction (DR) techniques cannot preserve all the structural characteristics of the original data. 
Therefore, DR techniques focus on preserving either local neighborhood structures (local techniques) or global structures such as pairwise distances between points (global techniques). However, both approaches can mislead analysts to erroneous conclusions about the overall arrangement of manifolds in HD data. For example, local techniques may exaggerate the compactness of individual manifolds, while global techniques may fail to separate clusters that are well-separated in the original space.
In this research, we provide a deeper insight into Uniform Manifold Approximation with Two-phase Optimization (UMATO), a DR technique that addresses this problem 
by effectively capturing local and global structures. 
UMATO achieves this by dividing the optimization process of UMAP into two phases. 
In the first phase, it constructs a skeletal layout using representative points, and in the second phase, it projects the remaining points while preserving the regional characteristics. 
Quantitative experiments validate that UMATO outperforms widely used DR techniques, including UMAP, in terms of global structure preservation, with a slight loss in local structure. 
We also confirm that UMATO outperforms baseline techniques in terms of scalability and stability against initialization and subsampling, making it more effective for reliable HD data analysis.
Finally, we present a case study and a qualitative demonstration that highlight UMATO’s effectiveness in generating faithful projections, enhancing the overall reliability of visual analytics using DR.

%% file: sections/01_introduction.tex
\section{Introduction}

\IEEEPARstart{D}{imensionality} reduction (DR) is a commonly used set of techniques to visualize high-dimensional (HD) data \cite{jo2018panene, fujiwara2019supporting, chatzimparmpas2020t} in various domains (e.g., bioinformatics \cite{becht2019dimensionality}, natural language processing \cite{boggust22iui}). 
DR techniques synthesize a low-dimensional representation (i.e., projection) that summarizes the structural characteristics of the original HD data, which can be visualized using scatterplots. 
As DR ``compresses'' data from a vast HD space to a narrow low-dimensional space, it cannot preserve all the structural characteristics of the original data. Therefore, each DR technique prioritizes preserving different structural characteristics.

%% local / global 동시에 만족해야 좋은 거를 보이는데 중점을 두자

In the literature, DR techniques can be broadly categorized into two groups---local techniques and global techniques---based on the structural characteristics they prioritize \cite{nonato2018multidimensional, xia2021revisiting, silva2003global}. 
Local techniques (e.g., UMAP \cite{mcinnes2018umap}, $t$-SNE \cite{maaten2008visualizing}) aim to preserve the local structures of HD data, such as neighborhood structures. In contrast, global techniques (e.g., PCA \cite{pearson1901liii}, Isomap \cite{tenenbaum2000global}, MDS \cite{kruskal64psycho}, and L-MDS \cite{de2004sparse}) \rev{focus on preserving large-scale relationships such as pairwise distances between distant points and the relative arrangement of manifolds, i.e., global structure.}

% focus on preserving the pairwise distances between data points to a broader scale, including those that are not neighbors. 

However, both local and global techniques fall short in generating projections that faithfully represent the arrangement of manifolds in the original HD data \cite{xia2021revisiting, jeon24tvcg, jeon2022measuring}. 
Local techniques ``exaggerate'' close neighbors while downplaying other relationships, resulting in projections that depict mutually more separated but individually more condensed manifolds (e.g., UMAP, Trimap, and PacMAP projections of Spheres data in \autoref{fig:syn_embeddings}).
For example, UMAP assumes that points that are not neighbors have no similarity \cite{mcinnes2018umap}.
\rev{This makes the resulting projections useful for identifying individual clusters and counting them, but not suitable for analyzing the distances between them.}
In contrast, global techniques often cause well-separated manifolds to overlap (e.g., PCA projection of S-Curve data in \autoref{fig:syn_embeddings}), \rev{potentially leading analysts to erroneous conclusions about the underlying structure.}
These errors can bias the perception of how the manifolds are arranged in the original dataset, resulting in an unreliable analysis of the data.
One way to alleviate this problem is to link multiple DR projections, for example, through small multiples or interactive methods \cite{jeon2022measuring, fujiwara23pvis}. 
However, this approach increases cognitive load on analysts. Moreover, static, non-interactive visualizations remain a common method for sharing data analysis results, as evidenced by their frequent use in many research papers across various fields \cite{cashman25tvcg}.

In this paper, we present UMATO, a DR technique \rev{designed to preserve both the global and the local structures}.
The main motivation \rev{of UMATO is to support users in both reliably identifying local manifolds (e.g., clusters or classes) while simultaneously examining their relationship (\autoref{sec:usecase}).} 
\rev{To achieve this, we} align UMATO with the typical visual analytics pipeline, which progresses from overview to detail \cite{shneiderman1996eyes}, by dividing the optimization into two sequential phases.  
In the first phase, optimization is performed on a small subset of representative points, i.e., hub points.
Since optimizing \rev{the distances between a small number of points} requires relatively less computation, 
the optimization considers the entire set of pairwise distances between points without any approximation.
Consequently, we obtain a skeletal layout that accurately preserves the global structure of the original data. 
In the second phase, we gradually add the remaining points to the projection. The resulting projection can accurately preserve the global structure because the aforementioned hub points are already embedded in place as anchors. 
In this phase, we employ the loss function and optimization procedure of UMAP to leverage its strength in accurately preserving local structures.

Our quantitative experiments show that UMATO achieves state-of-the-art performance in preserving the global structure while maintaining competitive performance in preserving local structures compared to other local DR techniques, which means that UMATO aligns well with our initial design goal.
Moreover, the scalability analysis shows that UMATO is faster than its competitors. 
Here, UMATO not only outperforms the original UMAP but also surpasses its faster variant algorithms, such as PacMAP \cite{wang21jmlr} and Trimap \cite{amid2019trimap}.
Additionally, we validate that UMATO is stable against noise in the data (e.g., subsampling) and substantially outperforms baseline techniques in this respect.
Lastly, a qualitative demonstration using four synthetic datasets and a case study reaffirm the capability of UMATO to faithfully represent the original HD data, leading to a more reliable data analysis.
Together, these results confirm the effectiveness of UMATO for reliable HD data analysis.

\boldsubsubsection{Improvements since the previous short paper}
Several enhancements have been made to the paper since we first introduce the core algorithm in our IEEE VIS 2022 short paper \cite{jeon22vis}. 
First, we enhance the algorithm for arranging outlier points to improve its accuracy (Appendix A).
Second, we conduct extensive evaluations of UMATO. While the previous short paper evaluates the accuracy of UMATO using three real-world datasets and a single synthetic dataset, we improve the generalizability of the accuracy evaluation by leveraging 20 real-world datasets.
We also present a case study demonstrating UMATO's effectiveness in supporting reliable visual analytics in real-world settings.
Moreover, we verify the effectiveness of UMATO in terms of stability and scalability.

We also improve the implementation of UMATO to facilitate its practical usage. 
First, we improve the scalability of the algorithm.
In our previous short paper, we report that UMATO is about three times slower than UMAP \cite{jeon22visappendix}. 
However, by optimizing the code to remove redundant calculations and parallelizing the algorithms, UMATO is now on par with UMAP. 
This also positions UMATO ahead of other state-of-the-art nonlinear DR techniques (\autoref{sec:scalexp}).
% Secondly, we illustrate how different hyperparameter values affect UMATO as guidance on setting hyperparameters in practice (\autoref{sec:hp}).
Finally, we make UMATO more accessible by offering it as an open-source Python library\footnote{\href{https://github.com/hyungkwonko/umato}{github.com/hyungkwonko/umato}}. 
As of \rev{August 2025}, UMATO has been downloaded over \rev{13,000 times}.

%% file: sections/02_related_works.tex
\section{Background and Related Work}

We discuss relevant literature in relation to our work. 
We first explain the UMAP algorithm in detail.
We then discuss two relevant areas: variants of UMAP and DR techniques for preserving global structure.

\subsection{UMAP}

\label{sec:umap}

UMATO adopts the loss function and optimization procedure of UMAP. 
We thus explain UMAP's computation procedure ($k$NN graph construction and layout optimization) in detail.
For the mathematical details, please refer to its original paper \cite{mcinnes2018umap}.

\boldsubsubsection{\textit{k}NN Graph Construction} After UMAP gets an HD data $X = \{x_1, \ldots, x_N\}$ as input, it constructs a weighted $k$NN graph.
Given $k$ (the number of NN to consider) and a distance function $d: X \times X \rightarrow [0, \infty)$, the $k$NN of $x_i$ regarding $d$, which we denote as $\mathcal{N}_i$, is computed.
Then, UMAP computes $\rho_i$, a distance from $x_i$ to its nearest neighbor:
\begin{equation} \label{eq:rho}
    \rho_{i} = \min_{j \in \mathcal{N}_i}\{ d(x_{i}, x_j) \ | \ d(x_{i}, x_j) > 0 \}.
\end{equation}
Subsequently, a parameter $\sigma_i$ satisfying:
\begin{equation} \label{eq:sigma}
    \sum_{j \in \mathcal{N}_i} \exp({-\max(0, d(x_{i}, x_j) - \rho_{i})} / \sigma_{i}) = \log_2(k).
\end{equation}
is found using a binary search. 
Next, UMAP computes the weight of the edge from $x_i$ to $x_j$, defined as:
\begin{equation} \label{eq:v}
    v_{j|i} = \exp({-\max(0, d(x_{i},x_{j})-\rho_{i})} / \sigma_{i}).
\end{equation}
A final weight of an edge connecting $x_i$ and $x_j$ is then defined as $v_{ij} = v_{j|i} + v_{i|j} - v_{j|i} \cdot v_{i|j}$.

\boldsubsubsection{Layout Optimization}
In this step, the algorithm aims to find a projection $Y = \{y_1, y_2, \cdots, y_N \}$ that minimizes the loss between HD edge weights and low-dimensional similarities.
Here, UMAP defines the similarity between two points $y_i$ and $y_j$ in the projection as 
\begin{equation} \label{eq:w}
w_{ij} = (1 + a|| y_{i} - y_{j} ||_{2}^{2b})^{-1}, 
\end{equation}
where $a$ and $b$ are user-steerable hyperparameters.
Setting $a$ and $b$ to 1 is the same as using Student's $t$-distribution.

Cross-entropy between the edge weights ($v_{ij}$) and low-dimensional similarity ($w_{ij}$) is used for the loss function: 
\begin{equation} \label{eq:ce}
    CE = \sum_{i \neq j} [v_{ij} \cdot \log({v_{ij}} / {w_{ij}}) - (1-v_{ij}) \cdot \log((1-v_{ij}) / (1-w_{ij}))].
\end{equation}
UMAP uses spectral embedding~\cite{belkin2002laplacian} to initialize $y_i$. Then, $y_i$ positions are iteratively optimized to minimize $CE$.
Given the output weight $w_{ij}$ as $1/(1+ad_{ij}^{2b})$, where $d_{ij}^{2b} = || y_i = y_j ||_{2}^{2b}$, the attractive gradient is:
\begin{equation} \label{eq:pos}
    \frac{CE}{y_i} ^{+} = \frac{-2abd_{ij}^{2(b-1)}}{1+ad_{ij}^{2b}} v_{ij} (y_i - y_j),
\end{equation}
and the repulsive gradient is:
\begin{equation} \label{eq:neg}
    \frac{CE}{y_i} ^{-} = \frac{2b}{(\epsilon + d_{ij}^{2})(1 + ad_{ij}^{2b})} (1 - v_{ij}) (y_i - y_j).
\end{equation}
Note that $\epsilon$ is a small hyperparameter added to prevent division by zero, and $d_{ij}$ is the Euclidean distance between $y_i$ and $y_j$.

During optimization, the negative sampling technique is leveraged \cite{mikolov2013distributed, tang2015line, tang2016visualizing} for acceleration. 
The sampling is done by first choosing a target edge $(i, j)$ and $M$ negative sample points for each epoch.
Then, $i$ and $j$ contribute to attractive forces, and points in $M$ contribute to repulsive forces, where the positions of $i$, $j$, and $M$ are updated.
The objective function regarding negative sampling is like this: 
\begin{equation}    \label{eq:eq9}
    \widetilde{CE} = \sum_{(i, j) \in E} v_{ij} (\log (w_{ij}) + \sum_{k=1}^{M} E_{{j_k} \sim P_{n}(j)} \gamma \log (1 - w_{ij_{k}})).
\end{equation}
Here, $\gamma$ is a hyperparameter that defines the weight of negative samples. 
$E_{{j_k} \sim  P_n(j)}$ denotes that $j_k$ is sampled from a noisy distribution $P_n(j) \propto deg_j^{3/4}$ \cite{mikolov2013distributed}, where $deg_j$ denotes the degree of point $j$.

\textit{Our contributions.}
According to the original paper that introduces UMAP, the cross-entropy loss function that leverages both attractive and repulsive gradients makes UMAP accurately capture both local and global structures \cite{mcinnes2018umap} (\autoref{eq:ce}, \ref{eq:pos}, and \ref{eq:neg}). However, due to the edge weight function that focuses on $k$NNs (\autoref{eq:v}) and the limited number of samples through negative sampling, UMAP often falls short in preserving the global structure in practice \cite{jeon24tvcg, kobak2019umap, xia2021revisiting}.

UMATO's two-phase optimization scheme allows it to effectively exploit the capability of UMAP to capture local and global structures. In the first stage, UMATO optimizes a smaller number of points (i.e., hub points) without negative sampling approximation. Therefore, the technique \textit{fully leverages} the capability of UMAP's optimization strategy to capture the global structure. Then, in the second stage, UMATO optimizes the remaining points as UMAP does to leverage its capability to preserve local structures. Our quantitative experiments (\autoref{sec:quantexp}),  qualitative demonstrations (\autoref{sec:qualexp}), and a case study (\autoref{sec:usecase}) confirm UMATO's ability to represent the original structure of high-dimensional data accurately.

\subsection{Reliable High-dimensional Data Analysis with Dimensionality Reduction}

Visual analytics should be reliable, i.e., decision-making or knowledge generation based on visualization should accurately reflect the original data characteristics. 
However, HD data analysis with DR can easily become unreliable as distortions occur when projecting data from a vast HD space to a narrow low-dimensional space \cite{jeon2022measuring, lespinats11cgf, jeon2022distortion}.

A common approach to mitigate unreliability is to measure the accuracy of DR projections and use those with good scores. Diverse quality metrics have been proposed for this purpose \cite{nonato2018multidimensional}. While local metrics (e.g., Trustworthiness \& Continuity \cite{venna2001neighborhood}, MRREs \cite{lee2007nonlinear}) aim to measure how well DR projections preserve the local structure, global metrics (e.g., KL Divergence \cite{hinton2002stochastic}, Stress \cite{kruskal64psycho}) evaluate the preservation of the global structure.
For example, DR benchmark studies \cite{atzberger24tvcg, espadoto2019towards} use these metrics to identify the best matching projection for a given data or visual analytics task. 
These metrics can also be used to optimize hyperparameters to achieve the best projection possible with a given DR technique \cite{Moor19Topological, espadoto2019towards}.

We can also enhance the reliability of DR-based visual analytics by using multiple projections simultaneously \cite{jeon2022measuring, jeon24tvcg, fujiwara23pvis}. 
By juxtaposing multiple projections that focus on different structural characteristics, analysts can gain a more comprehensive and reliable understanding of the original HD data. 
However, using multiple projections requires more screen space, and linking different projections is mentally demanding \cite{fujiwara2019supporting, jackle17vast}. 
Consequently, an alternative strategy to augment a DR projection has emerged. For example, some studies have proposed to visualize distortions in different parts of the projection using techniques such as heatmaps \cite{martins14cg} or Voronoi diagrams \cite{lespinats11cgf, aupetit07neurocomputing}.

\textit{Our contribution.}
Despite all these efforts in the visualization community, it is still common to use a static DR projection to analyze data and share results. For example, many research papers \cite{bai2021uibert, lee22arxiv, lim23chi} and visual analytics systems \cite{narechania22tvcg, hong22pvis, kahng18tvcg} present and describe their data using a single DR projection. In such situations, our research contributes to achieving a more reliable data analysis by introducing a DR technique that produces projections that accurately reflect the manifold structure in the HD data.

\subsection{Dimensionality Reduction Techniques for Preserving both Local and Global Structures}

\label{sec:localglobal}

It is important to note that our research is not the sole work focusing on DR projections that preserve local and global structures. 
One typical strategy is to design a loss function incorporating both local and global aspects of HD data.
% Isomap \cite{tenenbaum2000global}, for example, utilizes geodesic distances to account for global structure.
Topological autoencoder (TopoAE)  \cite{Moor19Topological}, for example, achieves the goal by adding a topological loss that considers global structure to the original reconstruction loss of autoencoders that makes the algorithm better preserve local structure \cite{hinton2006reducing}. 
Another approach is to modify the distance function. PacMAP \cite{wang21jmlr}, a variant of UMAP, introduces a flexible distance function that adapts based on the density of the data. TriMAP \cite{amid2019trimap}, another variant of UMAP, defines weights (i.e., similarity) between data points in triplets rather than pairs. However, these techniques optimize all points simultaneously, which means both global and local structures are optimized together. This approach can potentially compromise the preservation of either the local or global structures.

As an alternative, approaches using skeletal points have emerged. 
These points are often referred to as \textit{hubs}, \textit{landmarks}, or \textit{anchors}.
For example, De Silva and Tenebaum proposed L-Isomap \cite{silva2003global}, which extends the Isomap by leveraging landmarks. 
Joia et al. \cite{joia2011local} introduced LAMP, which allows users to steer projections by moving landmarks. 

\textit{Our contributions.}
However, these techniques randomly choose landmarks without considering their structural importance, resulting in an inaccurate representation of the global structure. 
Also, these techniques are designed by modifying DR techniques that perform suboptimally in preserving local structures, making accurate local structure preservation challenging.
In summary, these techniques hardly reach the full potential of targetting both local and global structure preservation.
In contrast, UMATO utilizes hubs (equivalent to landmarks) that are systematically chosen to better capture the global structure, achieving state-of-the-art performance in preserving the global structure of HD data (\autoref{sec:accuexp}).
Also, by leveraging UMAP's optimization procedure, a state-of-the-art algorithm for capturing local structure, UMATO accurately captures not only the global structure but also the local structure of HD data.

%% file: sections/03_umato.tex
\begin{figure*}[t]
    \centering
    \includegraphics[width=\linewidth]{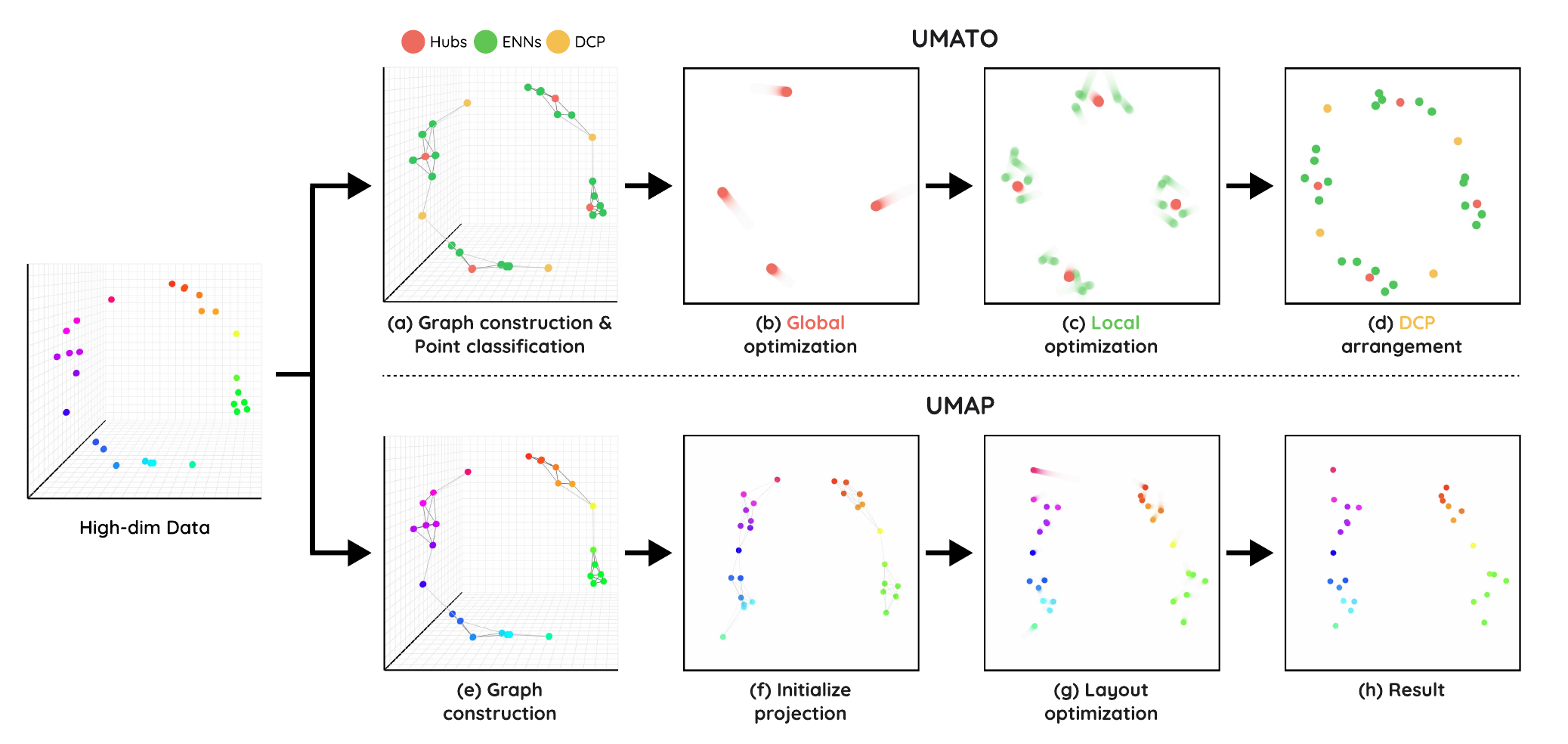}

    \vspace{-6mm}
    \caption{The comparison between the pipelines of UMAP and UMATO. 
    Based on a given HD data, UMATO first constructs a $k$NN graph and classifies points into three groups (hubs, extended nearest neighbors or eNNs, and disconnected points or DCPs) using the $k$NN indices (a).
    In the layout optimization stage, hubs, eNNs, and DCPs are embedded separately in order (b-d).  
    Note that UMATO also starts by initializing hubs, but we omit this in the figure.
    The separation of optimization enhances UMAP's stability and accuracy in preserving global structure. 
    In contrast, UMAP does not classify points and optimizes every point together, compromising its stability and precision in maintaining the global structure (e-h).}
    \label{fig:illust}
\end{figure*}

\begin{algorithm}[t]
\small
\caption{UMATO}
\label{alg:UMATO}
\begin{algorithmic}[1]
\Procedure{UMATO}{$X$, $k$, $d$, $n_{h}$, $e_{g}$, $e_{l}$}
    \Require High-dimensional data $X$, number of nearest neighbors $k$, projection dimension $d$, number of hub points $n_{h}$, epochs for global and local optimization $e_{g}$, $e_{l}$
    \Ensure Low-dimensional projection $Y$
    \State Compute $k$-nearest neighbors of $X$
    \State Obtain a sorted list using the indices' frequency of $k$-nearest neighbors
    \State Build $k$-nearest neighbor graph structure
    \State Classify points into hubs, expanded nearest neighbors, and disconnected points (\autoref{alg:classification})
    \State Optimize $CE(f(X_{h})||g(Y_{h}))$ to preserve global configuration (\autoref{eq:ce})
    \State Initialize expanded nearest neighbors using hub locations
    \State Update $k$-nearest neighbors \& compute weights (\autoref{eq:v})
    \State Optimize $\widetilde{CE}(f(X)||g(Y))$ to preserve local configuration (\autoref{eq:eq9})
    \State Arrange disconnected points
    \State \Return{$Y$}
\EndProcedure
\end{algorithmic}
\end{algorithm}

\section{UMATO}

\label{sec:algorithm}

\begin{algorithm}[t]
\small
\caption{Point Classification}
\label{alg:classification}
\begin{algorithmic}[1]
\Procedure{Point Classification}{$X$, $K$, $n_{h}$}
\Require High-dimensional data $X$, $k$-nearest neighbor indices $K$, number of hub points $n_{h}$
\Ensure Point classes $P_h$, $P_e$, $P_d$
\State $P_h = \emptyset$, $P_e = \emptyset$, $P_d = \emptyset$
\State $K_p = \{(x_i, f_i) | x_i \in \text{Flatten}(K) $ with $f_i$ being the corresponding frequency$\}$
\For{$i=1$ to $n_h$}
    \State $P_h \gets P_h \cup { x_i }$ where $x_i$ has the largest $f_i$ in $K_p$
    \State $K_p \gets K_p - \text{NN}_{1}{(x_{i})}$ where $\text{NN}_{1} = \{x_j | (x_j, f_j) \in K_p$ and $\forall x_j \in \text{NN}\}$
    % \State Calculate NNs of $x_i$ in $K_p$, where $\forall x_j \in \text{NNs}$ and $(x_j, f_j) \in K_p$. Remove them from $K_p$
\EndFor
\For{$i=1$ to $n_h$}
    \State $P_e \gets P_e \cup \text{NN}_{2}(x_i)$ where $x_i \in P_h$ and $\text{NN}_{2}(x_i)$ is the NNs of $x_i$ obtained from $K$
\EndFor
\State $P_d \gets X - (P_h \cup P_e)$
\State \Return{$P_h$, $P_e$, $P_d$}
\EndProcedure
\end{algorithmic}
\end{algorithm}

\begin{table*}[t]
\renewcommand{\arraystretch}{1.3}
    \centering
    
    \caption{The list of HD datasets used in the quantitative experiments (\autoref{sec:quantexp}) and their traits. For detailed explanations about the traits, please refer to Espadoto et al. \cite{espadoto2019towards}.}
    \scalebox{0.95}{
    \begin{tabular}{l|lllllllll}
    \toprule
       \textbf{Dataset}  & \textbf{Type} & \textbf{Size} & \textbf{Size class} & \textbf{Dim.} & \textbf{Dim. Class} & 
       \textbf{Int. Dim.} & \textbf{Int. Dim. Class} & \textbf{Sparsity} & \textbf{Sparsity Class} \\
       \midrule
       Blood Transfusion Service Center \cite{yeh09esa}  & Table & 748 & small & 4 & low & 0.2500 & medium & 0.0017 & dense \\ 
       Asteroseismology \cite{hon2017deep} & Table & 1001 & medium & 3 & low & 0.3333 & medium & 0.0000 & dense  \\ 
       CNAE-9 \cite{asuncion07uci} & Text & 1080 & medium & 856 & high & 0.3960 & medium & 0.9922 & sparse \\ 
       Coil-20 \cite{nene20columbia} & Image & 1440 & medium & 400 & medium & 0.2675 & medium & 0.3691 & medium \\ 
       \underline{Epileptic Seizure Recognition} \cite{andrzejak2001indications} & Table & 5750 & large & 178 & medium & 0.0291 &  medium &  0.0002 & dense \\
       \rev{Flickr} Material Database \cite{sharan2009material} & Image & 997 &      small & 1536 &      high &       0.3066 &              medium &        0.0010 &                dense \\
       \underline{Hate Speech} \cite{davidson2017automated} & Text & 3221 &      large &  100 &    medium &       0.8600 &                high &        0.9701 &               sparse \\ 
       \underline{IMDB} \cite{maas2011learning} & Text & 3250 &      large &  700 &      high &       0.8171 &                high &        0.9417 &               sparse \\ 
       Ionosphere \cite{asuncion07uci} & Table & 351 &      small &   34 &       low &       0.7058 &                high &        0.1191 &                dense \\ 
       MNIST64 \cite{xia2021revisiting} & Image & 1082 &     medium &   64 &       low &       0.4218 &              medium &        0.4935 &               medium \\ 
       \underline{Optical Recognition} \cite{asuncion07uci} & Image & 3823 &      large &   64 &       low &       0.4531 &              medium &        0.4880 &               medium \\
       \underline{Paris Housing}  \cite{kaggle}& Table & 10000 &      large &   17 &       low &       0.0588 &                 low &        0.1520 &                dense \\
       \underline{Predicting Pulsar Star} \cite{kaggle} &  Table  & 9273 & large        &     8 & low         &        0.2500     & medium                &      0.0000           & dense \\
       Raisin \cite{ccinar2020classification} & Table &  900 &      small &    7 &       low &       0.1429 &              medium &        0.0000 &                dense \\ 
       \underline{Rice Seed (Gonen Jasmine)} \cite{kaggle} & Table & 18185 &      large &   10 &       low &       0.1000 &                 low &        0.0000 &                dense \\ 
       Seismic Bumps \cite{sikora2010application} & Table & 646 &      small &   24 &       low &       0.2917 &              medium &        0.5827 &               medium \\
       Sentiment Labeled Sentences \cite{kotzias2015group} & Text & 2748 &     medium &  200 &    medium &       0.8800 &                high &        0.9887 &               sparse \\
       SMS Spam Collection \cite{almeida2011contributions} & Text & 835 &      small &  500 &      high &       0.6700 &                high &        0.9914 &               sparse \\
       Weather \cite{ventocilla2020comparative} & Table & 365 &      small &  192 &    medium &       0.06250 &                 low &        0.0033 &                dense \\
       Website Phishing \cite{abdelhamid2014phishing} & Table & 1353 &     medium &    9 &       low &       0.8888 &                high &        0.3199 &               medium \\
       \bottomrule
       \addlinespace[0.115cm]
       \multicolumn{10}{l}{
        \footnotesize
        (1) \textbf{Type}: The category to which a dataset belongs (\textit{Table}, \textit{Text}, or \textit{Image})
        } \\ 
        \multicolumn{10}{l}{
        \footnotesize
        (2) \textbf{Size}: Number of data points (i.e., samples) in a dataset (\textit{small}: $N \leq 1000$, \textit{medium}: $1000 < N \leq 3000$, \textit{large}: $N > 3000$)
        } \\ 
        \multicolumn{10}{l}{
        \footnotesize
        (3) \textbf{Dim.} (Dimensionality): Number of dimensions of a dataset (\textit{small}: $D < 100$, \textit{medium}: $100 \leq D < 500$, \textit{high}: $D \geq 500$)  

        } \\ 
       \multicolumn{10}{l}{
        \footnotesize
        (4) \textbf{Int. Dim.} (Intrinsic Dim.): The percentage of principal components needed to explain 95\% of the data variance 
        } \\ 
        \multicolumn{10}{l}{
        \footnotesize
        \hspace{36.5mm}
        (\textit{low}: $D_I \leq 0.1$, \textit{medium}: $0.1 < D_I \leq 0.5$, \textit{high}: $0.5 < D_I \leq 1$)
        } \\ 
        \multicolumn{10}{l}{
        \footnotesize
        (5) \textbf{Sparsity}: The ratio of non-zero values in a dataset (\textit{dense}: $S \leq 0.2$, \textit{medium}: $0.2 < S \leq 0.8$, \textit{sparse}: $0.8 < S \leq 1$)  
        } \\

    \end{tabular}
    }
    \label{tab:datasets}
\end{table*}

We introduce UMATO, a DR algorithm for more reliable analysis of HD data manifolds. 
Aligned with Shneiderman's visual information-seeking mantra \cite{shneiderman1996eyes},
\textit{Overview first, zoom and filter, and details on demand,}
\rev{UMATO first projects skeletal points to} \rev{preserve the global structure, then projects the remaining points while focusing on local structure preservation.}
\rev{By doing so, UMATO helps users to reliably identify local manifolds and examine their relationship.}
Please refer to \autoref{fig:illust}, \autoref{alg:UMATO}, and \autoref{alg:classification} for detailed illustrations of the algorithm.

% improving UMAP by dividing its optimization into two phases. 
% UMATO enhances UMAP in terms of accuracy in global structure preservation, stability, and scalability.

% We present UMATO, which splits UMAP's optimization into two phases to \rev{preserve global structure better while maintaining the capability to represent local structure}. 
% For ease of understanding, we illustrated the UMATO pipeline in \autoref{fig:illust}, presented the pseudo code in \autoref{alg:UMATO}, and .

\subsection{\textit{k}NN Graph Consturction}

\label{sec:knnconst}

UMATO shares the initial step with UMAP. 
We first construct $k$NN indices. 
Then, by calculating  $\rho_i$ (\autoref{eq:rho}) and $\sigma_i$ (\autoref{eq:sigma}) for each point $i$, we obtain the pairwise similarity for every pair of points in $k$NN indices.

\subsection{Point Classification} 

\label{sec:ptclassi}

%% TODO

\rev{
The objective of UMATO is to enable users to reliably identify local manifolds and investigate their relationship. 
To this end, UMATO}
classifies the points into three disjoint sets---hubs ($P_h$), expanded nearest neighbors (eNNs or $P_e$), and disconnected points (DCPs or $P_d$).
\rev{The role of hubs is to establish the skeletal layout that represents the global structure of the input data. Hubs are distributed proportionally to density, anchoring the data in a manner that accurately preserves the global relationships between important local manifolds (e.g., dense clusters). The eNNs and DCPs are then projected with the objective of precisely depicting the local structure within such manifolds.
}

\rev{The procedure of point classification is as follows:}
we first calculate how many times each point appears as a $k$NN of other points, i.e., the frequency of each point in the $k$NN indices.
We then make a sorted list of points in descending order based on their frequency.
Next, we iteratively run the following two steps until all points are connected: 1) designate the point with the highest frequency as a hub from the pool of points that have not been selected yet; 2) remove $k$NN of the selected hub from the sorted list.
By using the sorted list, we make the hub picked in each iteration to be both popular and sufficiently dispersed from other hubs that have already been chosen.
\rev{Hubs can thus be interpreted as mutually dissimilar points with high local density \cite{zhao25tvcg}. Such points are widely recognized as carrying crucial information for approximating the original structure of data \cite{hou20pr, rodriguez14science}, thereby justifying our design choice.}
Once these hubs and their $k$NN are set, we 
recursively identify $k$NN of the current $k$NN until no additional points can be appended.
These recursively identified neighbors, except for the hubs, are referred to as eNN.
Any set of points not belonging to either hubs or eNNs is classified as DCPs.
Such points occur because their NNs are located far away, thus having another set of points as their NNs.

\subsection{Layout Optimization}

\label{sec:layoutopt}

We take different strategies to optimize different sets of points.
This is to improve the preservation of both the global and local structures of the data.
After capturing the global structure using only the hubs, we capture the local structure by embedding the eNNs.
We refrain from optimizing DCPs, as it has been observed to potentially corrupt the overall arrangement of manifolds.

\label{sec:layout}

\boldsubsubsection{Global Optimization}
To build the skeletal layout of the projection, we run the global optimization for the hubs.
We start by using PCA to set the initial positions of hub points.
We use PCA because it has been verified to support the final projection in better capturing global structure \cite{kobak2021initialization}.
Moreover, PCA is substantially faster than UMAP's initialization method (Spectral embedding), thus enhancing the overall scalability of UMATO (\autoref{sec:scalexp}).

Then, we optimize their positions by minimizing the cross-entropy function (\autoref{eq:ce})
Specifically, let $f(X) = \{ f(x_{i}, x_{j}) | x_{i}, x_{j} \in X \}$ and $g(Y) = \{ g(y_{i}, y_{j}) | y_{i}, y_{j} \in Y \}$ be two adjacency matrices in high- and low-dimensional spaces, respectively.
Then, $CE(f(X_{h}) || g(Y_{h}))$ is minimized, where $X_{h}$ represents a set of points selected as hubs in HD space and $Y_{h}$ is a set of corresponding points in the projection.
The global optimization process does not include negative sampling approximation, which makes the projection more robust and less biased in capturing global structure.
Moreover, it requires relatively less time since it runs only for the selected hub points.

\boldsubsubsection{Local Optimization}
Next, UMATO embeds \rev{eNNs}, mainly aiming to capture local structure.
We set the initial position of each data point $x \in X$ \rev{in the 2D projection} as an average position of $m$ (e.g., 10) NN with a small random perturbation.
UMAP's optimization starts by building a $k$NN graph (see \autoref{sec:umap}); we conduct the same task but only with $x_i \in P_h \cup P_e$.
To this end, we update $k$NN indices constructed in advance (\autoref{sec:ptclassi}) to rule out the DCPs.
In detail, regarding any point $x_i$ in the set $P_h \cup P_e$ and its neighbors $x_{i_{j}} \in N_{x_i}$ (where $1 \leq j \leq k$), if $x_{i_{j}}$ belongs to the set $P_d$, we exclude it from $N_{x_i}$ and update it as the next neighbor $x_{i_{k+1}}$, ensuring that $x_{i_{k+1}} \notin P_d$.
Since we use the $k$NN indices we have already built, the computation is not expensive. 

Afterward, local optimizations of hubs and eNNs (i.e., $x_i \in P_h \cup P_e$) are performed based on the cross-entropy loss function, similar to UMAP. We also leverage the negative sampling technique (\autoref{eq:eq9}).
However, UMATO prioritizes preserving the positions of hubs due to their established role in the global structure, favoring this approach over uniform updates of all points' positions.
We achieve this by selecting $i$ among eNNs and choosing $j$ from both hubs and eNNs to sample a target edge $(i, j)$. 
If $j$ is a hub, we penalize the attractive force for $j$ by assigning a small weight (e.g., 0.1), which makes $j$ not excessively affected by $i$ if it is a hub point.
Furthermore, the repulsive force can disperse local attachments, causing points to deviate in each epoch and ultimately disrupting the well-structured global layout.
To mitigate this, we consider a penalty (e.g., 0.1) when calculating the repulsive gradient (\autoref{eq:neg}) for the points selected as negative samples.

\boldsubsubsection{Disconnected Points Arrangement}
Unlike hubs or eNNs, DCPs are almost equidistant from all the other data points in HD space because of the curse of dimensionality \cite{bellman1966dynamic, lee2007nonlinear}.
Incorporating them into the optimization can make them mingle with the already positioned points (i.e., hubs, eNNs), potentially disrupting both global and local structures.
We thus project DCPs near their NNs; for a DCP $x_i \in P_d$, we embed $x_i$ on the centroid of $k$NNs of $x_i$.
This approach allows us to benefit from the overall composition of the already optimized projection.

\subsection{Computational Complexity}

\label{sec:complexity}

We analyze the time complexity of optimizing UMATO as follows. 

\boldsubsubsection{\textit{k}NN graph construction (\autoref{sec:knnconst})} As with UMAP, constructing $k$NN indices relies on the Nearest-Neighbor-Descent algorithm \cite{dong11www}, which costs $O(N^{1.14})$, where $N$ stands for the size of the dataset. 

\boldsubsubsection{Point Classification (\autoref{sec:ptclassi})} We \rev{sort the points and} visit each point once while classifying points. Therefore, the time complexity is  \rev{$O(N
\log N)$}.

\boldsubsubsection{Layout Optimization (\autoref{sec:layout})} 
Regarding global optimization, PCA initialization on hub points requires \rev{$O(d|P_h|)$}. Each epoch of optimization costs \rev{$O(|P_h|^2)$} as the stage runs without negative sampling approximation. 
Meanwhile, each epoch of the local optimization costs, \rev{which incorporates negative sampling, is $O(k * (|P_h| + |P_e|))$} as the attractive forces need to be calculated for all neighbor edges. \cite{mcinnes2018umap}. As each DCP can be embedded in constant time, the time complexity of the DCP arrangement step is $O(|P_d|)$. 

Combining these, the overall time complexity of UMATO optimization is \rev{$O(N^{1.14} + N log N + d|P_h| + |P_h| ^ 2 + k (|P_h| + |P_e|) + k |P_d|)$, which can be further simplified to $O(N ^ {1.14} + |P_h| ^ 2 + kN)$}. The complexity is slightly higher than UMAP (which is \rev{$O(N^{1.14} + kN)$} \cite{mcinnes2018umap}), 
% However, our implementation of UMATO is faster than UMAP as it uses a much faster initialization method, specifically PCA. 
\rev{making UMATO marginally slower than UMAP when using PCA initialization (\autoref{sec:scalexp}).}

%% file: sections/04_implementation.tex
\sethlcolor{Apricot}

\newcommand{\f}[1]{\hl{ \textbf{{\underline{#1}}} }}
\newcommand{\s}[1]{\textbf{\underline{#1}}}
\newcommand{\h}[1]{\textbf{#1}}

\newcommand{\pzero}{\cellcolor{blue!48}}
\newcommand{\pone}{\cellcolor{blue!40}}
\newcommand{\ptwo}{\cellcolor{blue!32}}
\newcommand{\pthree}{\cellcolor{blue!24}}
\newcommand{\pfour}{\cellcolor{blue!16}}
\newcommand{\pfive}{\cellcolor{blue!8}}
\newcommand{\psix}{\cellcolor{white}}
\newcommand{\pseven}{\cellcolor{red!8}}
\newcommand{\peight}{\cellcolor{red!16}}
\newcommand{\pnine}{\cellcolor{red!24}}
\newcommand{\pten}{\cellcolor{red!32}}
\newcommand{\peleven}{\cellcolor{red!40}}
\newcommand{\ptwelve}{\cellcolor{red!48}}

\newcommand{\xone}{\cellcolor{blue!48}}
\newcommand{\xtwo}{\cellcolor{blue!32}}
\newcommand{\xthree}{\cellcolor{blue!16}}
\newcommand{\xfour}{\cellcolor{red!16}}
\newcommand{\xfive}{\cellcolor{red!32}}
\newcommand{\xsix}{\cellcolor{red!48}}

\begin{table*}[t]
\renewcommand{\arraystretch}{1.3}
    \centering
    \scriptsize
    \caption{The average scores that 13 DR techniques obtain in our first accuracy analysis (\autoref{sec:compprac}).
    For each quality metric, DR techniques ranked between first and sixth place are highlighted in \colorbox{blue!25}{blue}, where we assign higher opacity to the better techniques. Similarly, techniques ranked between eighth and thirteenth place are highlighted in \colorbox{red!25}{red}, where worse techniques have higher opacity. 
    UMATO substantially outperforms the baselines in terms of global metrics with a slight sacrifice in local metric scores. Note that we standardize both the original data and projections to minimize the impact of scaling \cite{smelser24beliv}.
    }
    \scalebox{0.92}{
    \begin{tabular}{l cccccccc cc ccccc}
        \toprule
         & \multicolumn{10}{c}{Local}  & \multicolumn{5}{c}{Global}  \\
         \cmidrule(lr){2-11} \cmidrule(lr){10-11} \cmidrule(lr){12-16}
         & \makecell{Trust. \\ $k=10$} & \makecell{Trust. \\ $k=50$} & \makecell{Conti. \\ $k=10$} & \makecell{Conti. \\$k=50$} & \makecell{MRRE$_F$ \\ $k=10$} & \makecell{MRRE$_F$ \\ $k=50$} & \makecell{MRRE$_M$ \\ $k=10$} & \makecell{MRRE$_M$ \\ $k=50$} & Stead. & Cohev. & \makecell{KL Div. \\ $\sigma=1$} & \makecell{KL Div. \\ $\sigma=.1$} & \makecell{DTM \\ $\sigma=1$} & \makecell{DTM \\ $\sigma=.1$} & Stress\\
         \midrule
         UMAP           & \pthree 0.9067 & 0.8658  \pfour   & \pone 0.9420 & 0.8773  \peight   & \pthree 0.9113 & \pthree 0.8922 & \pone 0.9524 & \pone 0.9227  & \ptwo 0.8538 & \ptwo 0.6445  & 0.0042  \ptwelve  & 0.2383  \ptwelve   & 0.0662  \ptwelve   & 0.4056  \ptwelve   & 2.7369 \peight \\
        UMAP (PCA) & \ptwo 0.9086 & \pthree 0.8675 & \ptwo 0.9413 & \pfive 0.8885 & \ptwo 0.9137 & \ptwo 0.8943 & \ptwo 0.9526 & \ptwo 0.9267 & \pthree 0.8491 & \pone 0.6510 & \pten 0.0034 & \peleven 0.2005 & \peleven 0.0579 & \peleven 0.3852 & \pnine 2.7735 \\
         $t$-SNE        & \pzero 0.9218  & \pone 0.8727  & \pzero 0.9442  & \pzero 0.9049 & 0.9327 \pzero & \pzero 0.9087 & \pzero 0.9561 & \pzero 0.9351 & \pzero 0.8605 & 0.6066  \peight   & 0.0030   \peight  & \pthree 0.1445 & 0.0581  \pten   & 0.3717   \peight  &  7.4736 \ptwelve \\
         LLE            & 0.8495  \peight   & 0.8300 \pten    & 0.9116  \pseven    & 0.8790 \pseven    & 0.8515   \peight  & 0.8398  \peight  & 0.9202  \pseven    & 0.9012 \pseven     & 0.7459  \peight   & \pfive 0.6226 & 0.0042  \ptwelve   & 0.1905  \pnine   & 0.0550   \peight & 0.3775  \pten   &  0.9909 \pseven\\
         PacMAP         & \pone 0.9194  & \pzero 0.8869  & \pfive 0.9227 & \psix 0.8862 & \pone 0.9225 & \pone 0.9067 & \pfive 0.9293 & \pfive 0.9111  & \pone 0.8557  & 0.5999   \pten  & 0.0026 \psix    & 0.1521  \pfive   & 0.0517  \psix   & 0.3429  \psix   & 4.5020 \peleven\\
         Trimap         & \psix 0.8954     & \pfive 0.8705  & 0.8891  \pnine   & 0.8524  \pten   & 0.8987  \psix   & 0.8851 \psix    & 0.9025  \pnine  & 0.8805  \pnine   & 0.8510  \pfour   & 0.6221  \psix   & 0.0030   \peight  & 0.1899  \peight   & 0.0546  \pseven   & 0.3819 \peleven     & 3.1781 \pten\\
         LAMP           & 0.7482   \ptwelve  & 0.7360   \ptwelve  & 0.8759  \peleven   & 0.8277  \ptwelve   & 0.7535  \ptwelve   & 0.7432  \ptwelve   & 0.8940  \peleven   & 0.8657  \peleven   & 0.5104  \ptwelve   & 0.5342 \ptwelve    & \pfive 0.0021  & \ptwo 0.1306  & \pfive 0.0418  & \pfive 0.3167  & \ptwo 0.6359  \\
         L-MDS          & 0.8339  \peleven   & 0.8254 \peleven    & 0.8685   \ptwelve  & 0.8393   \peleven  & 0.8374  \peleven   & 0.8290 \peleven    & 0.8815  \ptwelve   & 0.8616   \ptwelve  & 0.7039  \peleven   & 0.5989 \peleven    & 0.0039   \peleven  & 0.1986 \pten    & 0.0591 \peleven    & 0.3759  \pnine   & 0.9521 \psix\\
         PCA            &0.8406  \pten    & 0.8367 \pnine   & 0.9006  \peight   & \ptwo 0.8902 & 0.8431  \pten   & 0.8371  \pten   & 0.9074  \peight   & 0.8972  \peight   & 0.7288  \pten   & \pfour 0.6362 & \pthree 0.0020 & 0.1681  \pseven  & \pthree 0.0369 & \pfour 0.3114 & \pone 0.4362  \\
         Isomap & 0.8560 \peight & 0.8437 \peight & 0.9282 \pthree & 0.8983 \pthree & 0.8595 \peight & 0.8503 \peight & 0.9360 \pthree& 0.9187 \pthree& 0.7812 \psix & 0.6735 \pone & 0.0021 \pfive & 0.1536 \psix & 0.0376 \pfour & 0.2979 \pthree & 0.8468 \pfive \\ 
         MDS & 0.8370 \ptwelve & 0.8414 \pnine & 0.8936 \peight & 0.8976 \pfour & 0.8373 \ptwelve & 0.8350 \pnine & 0.8938 \pten & 0.8914 \psix & 0.7712 \pnine & 0.6772 \pzero & 0.0004 \pzero & 0.0823 \pzero & 0.0135 \pzero & 0.2070 \pzero & 0.2193 \pzero \\ 
         \textbf{UMATO} (Rand.) & 0.8619 \pseven & 0.8399 \peleven & 0.9180 \psix & 0.8811 \pseven & 0.8650 \pseven & 0.8522 \pseven & 0.9231 \psix & 0.9041 \psix & 0.7805 \pseven & 0.5847 \peleven & 0.0019 \ptwo & 0.1418 \pthree & 0.0372 \ptwo & 0.3118 \pfour & 0.8334 \pfour \\
         \textbf{UMATO} & 0.8716 \pfive & 0.8527 \psix & 0.9266 \pfour & 0.8989 \pone & 0.8747 \pfive & 0.8627 \pfive & 0.9303 \pfour & 0.9150 \pfour & 0.7716 \peight & 0.6178 \pseven & 0.0015 \pone & 0.1290 \pone& 0.0348 \pone & 0.2915 \ptwo & 0.8391 \pthree \\
         \bottomrule
    \end{tabular}
    }
    \label{tab:accuracy}
\end{table*}

%% file: sections/05_experiments.tex
\section{Quantitative Experiments}
\label{sec:quantexp}

We conduct a series of experiments to evaluate UMATO and compare it against competitors.
First, in \autoref{sec:accuexp}, we evaluate the accuracy of UMATO in depicting local and global structures of HD data. We then assess its scalability in \autoref{sec:scalexp}. Finally, in \autoref{sec:stabexp}, we examine the stability of UMATO against the subsampling and initialization methods. The experimental settings shared across all experiments are as follows.

\boldsubsubsection{Competitors}
Our key considerations in selecting competitors are as follows: 
(1) Competitors should be implemented in Python; \rev{we set this requirement to ensure that competitors are easily usable by data analysts in practice.} (2) Competitors should include global techniques, local techniques, and the ones that focus on both structures (referred to as \textit{hybrid techniques} for simplicity; \autoref{sec:localglobal}).
Based on these considerations, we select three local DR techniques (UMAP, $t$-SNE, LLE \cite{roweis00science}), four global techniques (PCA, Isomap \cite{tenenbaum2000global}, MDS \cite{kruskal64psycho}, and L-MDS \cite{de2004sparse}), and three hybrid techniques (LAMP \cite{joia2011local}, PacMAP \cite{wang21jmlr}, and Trimap \cite{amid2019trimap}).
For UMAP, PacMAP, Trimap, and LAMP, we use the implementation provided by the authors, which also leverages multithreading-based parallelization and thus can be fairly compared with our implementation. 
For $t$-SNE, we use the Multicore-TSNE library, and for PCA and Isomap, we use the scikit-learn \cite{pedregosa11jmlr} implementation. These two libraries also accelerate the techniques using multithreading.
For L-MDS, we use the implementation provided by Motta \cite{motta23github}.

\rev{To investigate the impact of initialization on performance, we add UMATO with random initialization instead of PCA (denoted as UMATO (random)) as a competitor.
We also include UMAP with PCA initialization (i.e., UMAP (PCA)) as a baseline to isolate and examine the effectiveness of UMATO's core algorithm beyond initialization (\autoref{sec:algorithm}).
}

\boldsubsubsection{Datasets}
We collect 20 HD datasets.
To ensure the diversity of datasets, we gather datasets with various traits (data type, size, dimensionality, intrinsic dimensionality, and sparsity), following the trait taxonomy proposed by Espadoto et al. \cite{espadoto2019towards}.
As a result, we construct a set of datasets that fully covers the taxonomy.
Please refer to \autoref{tab:datasets} for the list of datasets and their trait values.

% To ensure the diversity of datasets, we gather the datasets that have varying dataset traits (data type, size, dimensionality, intrinsic dimensionality, and sparsity), following 

\subsection{Accuracy Analysis}

\label{sec:accuexp}

\rev{We conduct two experiments that evaluate the accuracy of UMATO in preserving the structure of the original HD data. First, to assess the practical applicability of UMATO, we compare UMATO with aforementioned competitors that are likely to be used in practice (\autoref{sec:compprac}). Next, we compare UMATO against diverse variants of UMAP (e.g., the one that works without negative sampling) to provide an in-depth investigation into the effectiveness of our UMATO design (\autoref{sec:compvar}).}

\begin{figure}[t]
    \centering
    \includegraphics[width=\linewidth]{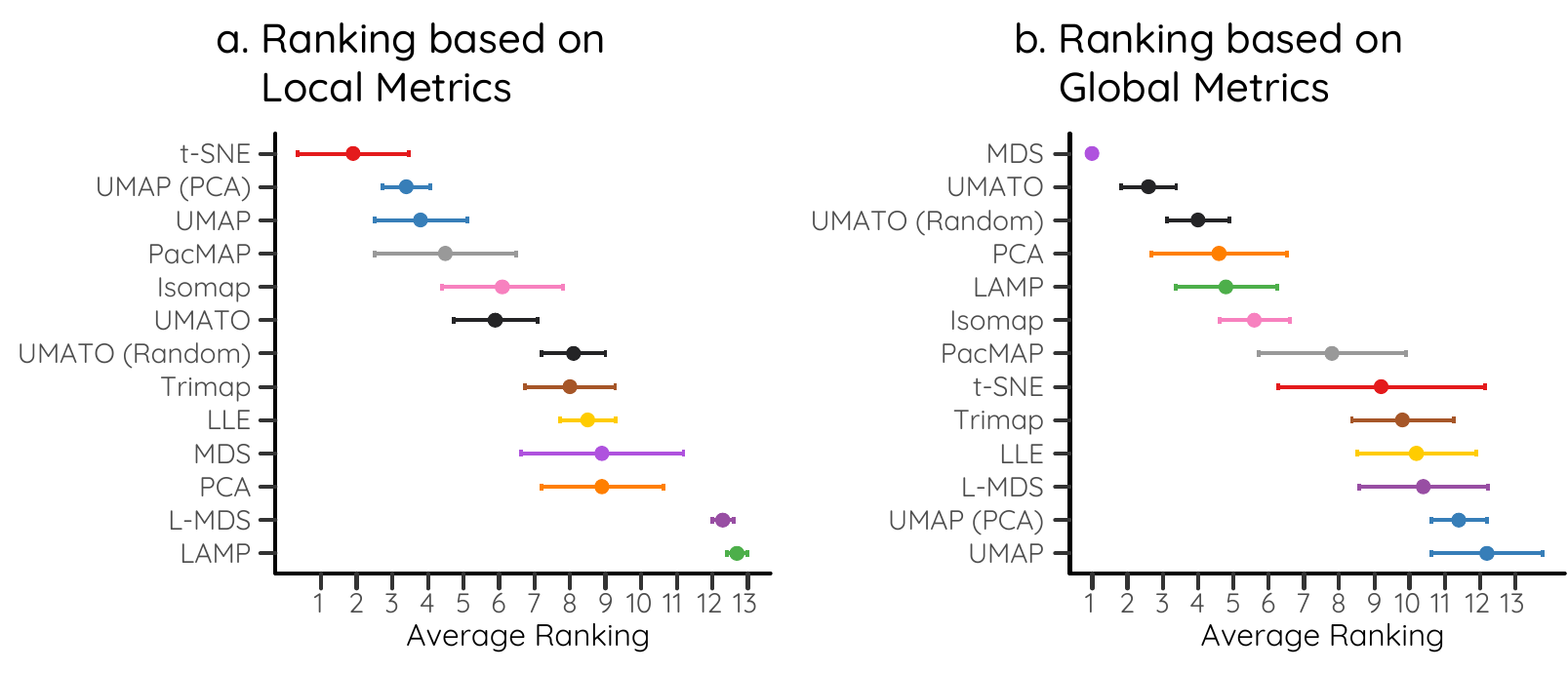}
    \vspace{-5mm}
    \caption{DR techniques ranked by local (a) and global (b) quality metrics in accuracy analysis (\autoref{sec:accuexp}, \autoref{tab:accuracy}). Among the ten techniques we compared, UMATO demonstrated the highest accuracy in terms of global metrics and showed intermediate performance for local metrics. 
    The error bars depict 95\% confidence intervals.
    Please refer to \autoref{tab:accuracy} for the detailed statistics.}
    \label{fig:rank_accuracy}
    % \vspace{-mm}
\end{figure}

\begin{figure*}[h]
    \centering
    \includegraphics[width=\textwidth]{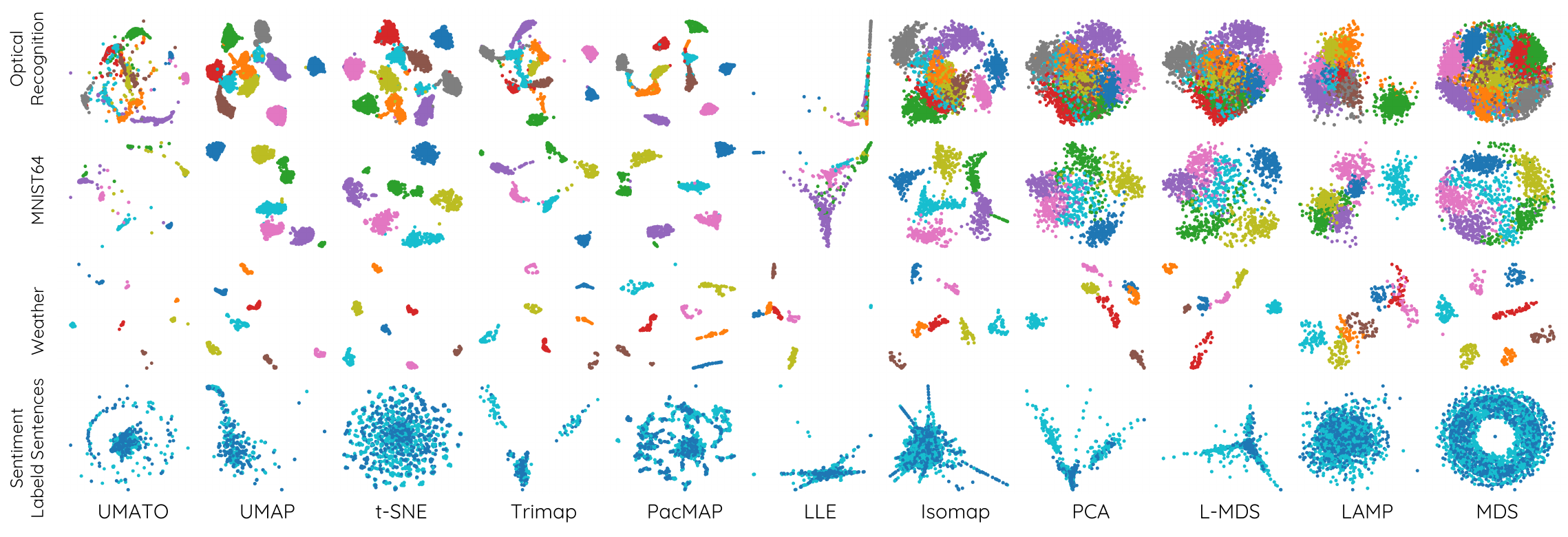}
    \vspace{-7mm}
    \caption{The subset of the projections generated in our accuracy analysis (\autoref{sec:accuexp}). Colors depict the class label of each dataset. The analysis results verified that UMATO outperforms competitors in terms of accurately preserving global structure while maintaining competitive performance in depicting local structure. Note that we only depict the projections made by default configurations for UMATO and UMAP.}
    \label{fig:accuracy_embeddings}
\end{figure*}

\begin{figure}[t]
    \centering
    \includegraphics[width=\linewidth]{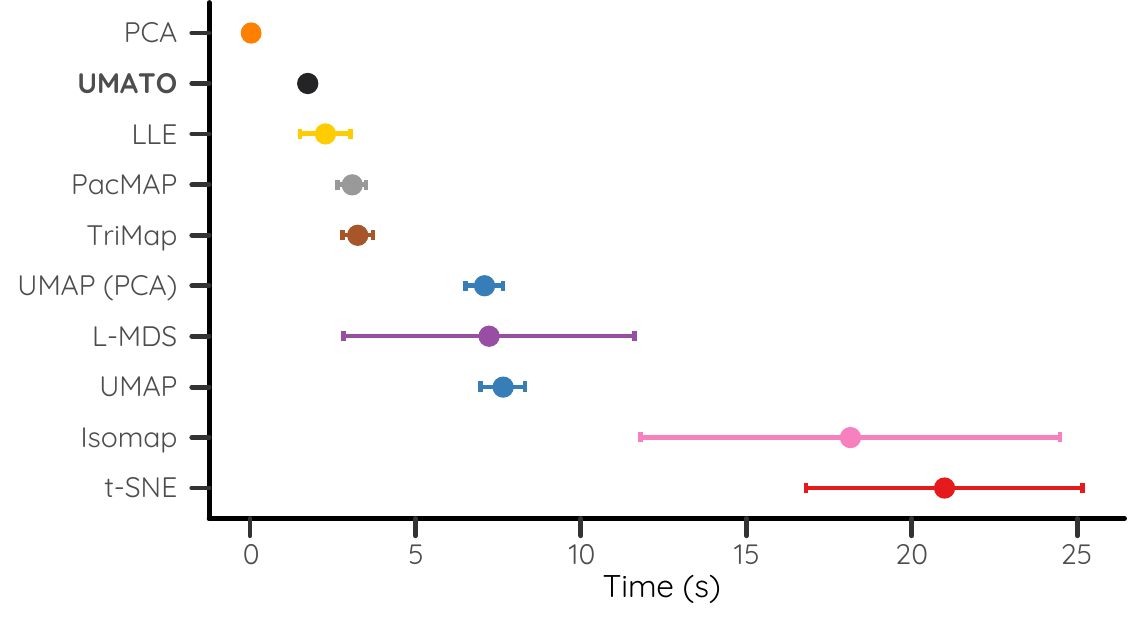}
    \vspace{-7mm}
    \caption{The results of the scalability analysis with small datasets (\autoref{sec:scalsmall}). Note that LAMP \rev{and MDS} have been removed from the figure as they need substantially longer computation time, making the runtime of all other techniques look similar.
    UMATO takes about three seconds on average to generate projections, outperforming all other nonlinear DR techniques. The error bars depict confidence intervals (95\%).}
    \label{fig:scal}
\end{figure}

\subsubsection{\rev{Comparison Against Practical Competitors}}

\label{sec:compprac}

\boldsubsubsection{Objectives and design}
We aim to evaluate the accuracy of UMATO, i.e., how accurately UMATO can preserve the global and local structures of the original HD data.
We generate the projections using UMATO and competitors, then assess accuracy using widely used local and global DR quality metrics.

% To evaluate accuracy in detail, we exploit diverse quality metrics from three categories that differ in target structural granularity: local, semi-global (i.e., cluster-level), and global \cite{jeon23vis}. 

\boldsubsubsection{Quality metrics}
We select the quality metrics from the list of representative metrics provided by Jeon et al. \cite{jeon23vis}. 
For local metrics, we use Trustworthiness \& Continuity (T\&C) \cite{venna2001neighborhood} and Mean Relative Rank Errors (MRREs) \cite{lee2007nonlinear}. 
Both metrics examine the extent to which $k$-nearest neighbor structure of the original and embedded spaces vary.
They thus require $k$ as a hyperparameter. 
Smaller $k$ forces the metrics to focus more on fine-grained local structure. We use two $k$ values, 10 and 50, for both metrics.
Note that as higher $k$ values make local metrics more focused on global structure, using two different values enhances the generalizability of our evaluation.
We also use Steadiness \& Cohesiveness (S\&C) \cite{jeon2022measuring} as a measure for examining the preservation of cluster structure. S\&C works by iteratively extracting clusters in one space and checking their dispersion in the other space. We classify S\&C as local metrics as it does not take into account the global arrangement of clusters by design \cite{jeon2022measuring}.
We use the default hyperparameter setting provided in the original paper.  

For global metrics, we use Kullback-Leibler (KL) Divergence  \cite{hinton2002stochastic}, Distance-to-Measure (DTM) \cite{chazal2011geometric}, and Stress \cite{kruskal64psycho}. 
KL divergence and DTM evaluate how accurately projections capture global structure in terms of density estimation, while Stress assesses this in terms of pairwise distances.
Both KL divergence and DTM require a hyperparameter $\sigma$, with higher values making the metrics focus more on the global structure. Following a previous convention \cite{Moor19Topological}, we use $0.1$ and $1$ as $\sigma$ value. 

\boldsubsubsection{Detailed procedure}
Following Moor et al. \cite{Moor19Topological}, 
we first generate optimal DR projections of datasets using Bayesian optimization \cite{snoek12nips}. 
We apply optimization to all DR techniques (UMATO and competitors), where the hyperparameter range we use is depicted in Appendix B.
We then evaluate the projections using quality metrics. F1 score of T\&C ($k=10$) is used as an optimization target, as T\&C is widely interpreted as precision and recall of DR \cite{lee2009quality, venna2010information}. \rev{Note that we replicate the experiment using global metrics (KL divergence) as target function in Appendix A, which shows consistent results.}
% Please refer to Appendix XX for the hyperparameter range we used.

\boldsubsubsection{Results and Discussions}
\autoref{tab:accuracy} depicts the detailed statistics of the experiment, and \autoref{fig:rank_accuracy} shows the overall ranking of techniques. 
\autoref{fig:accuracy_embeddings} shows the subset of projections generated in this experiment.

Regarding local metrics (T\&C, MRREs, S\&C), $t$-SNE shows the best performance, ranking first in eight out of ten metrics. UMAP and PacMAP are the runner-ups. 
Meanwhile, UMATO achieves intermediate accuracy, outperforming all global and hybrid techniques except PacMAP. Notably, UMATO even achieves a substantially better average ranking on local measures than LLE, a well-known local technique. 
In terms of global metrics, UMATO is \rev{one of the best techniques. MDS, the technique that directly optimizes the global distance, shows the best accuracy with UMATO and UMATO (random) as close runner-ups. Other DR techniques, including global techniques such as Isomap or PCA, perform worse overall than these techniques.} \rev{We also observe that PCA initialization improves UMATO, while it provides negligible benefit for UMAP. This finding demonstrates that preserving the global structure of HD data cannot be achieved through PCA initialization alone, yet the initialization still provides substantial benefit when combined with an effective optimization process. }

% Out of five metrics, UMATO ranks first in four metrics and third in the remaining metrics. PCA and Isomap follow UMATO. 

It is worth noting that UMAP shows the \textit{worst} accuracy in preserving global structure. This indicates that pairwise distances between distant points cannot be trusted in UMAP \cite{Coenen2019, jeon24tvcg}.
According to the \textit{Gestalt law on proximity}, which suggests that elements close to each other are perceived as related, this limitation can substantially undermine the reliability of visual analytics using UMAP.

In summary, UMATO is effective in preserving global structure while slightly sacrificing the capability to preserve the structure of local manifolds. 
This result aligns well with the design of UMATO: hubs help UMATO capture global structure in the first phase, but act as a constraint for local optimization of eNNs. 
These results clearly verify that UMATO can help analysts conduct HD data analysis in a more reliable manner. Meanwhile, the results again highlight the fact that accurately preserving both local and global structures can hardly be achieved.

% its merit in analyzing HD data.

\begin{table*}[t]
\renewcommand{\arraystretch}{1.3}
    \centering
    \scriptsize
    \caption{\rev{The average accuracy scores obtained by UMATO, UMAP and its variants (\autoref{sec:compvar}).
    DR techniques ranked between first and third place are highlighted in \colorbox{blue!25}{blue}, where we assign higher opacity to the techniques ranked ahead. Similarly, techniques ranked between fourth and sixth place are highlighted in \colorbox{red!25}{red}, where techniques that are ranked behind have higher opacity. The results show that turning off negative sampling results in worse accuracy of UMAP; such results support the effectiveness of our two-phase optimization design in preserving the global structure of the HD data. }
    }
    \scalebox{0.90}{
    \begin{tabular}{l cccccccc cc ccccc}
        \toprule
         & \multicolumn{10}{c}{Local}  & \multicolumn{5}{c}{Global}  \\
         \cmidrule(lr){2-11} \cmidrule(lr){10-11} \cmidrule(lr){12-16}
         & \makecell{Trust. \\ $k=10$} & \makecell{Trust. \\ $k=50$} & \makecell{Conti. \\ $k=10$} & \makecell{Conti. \\$k=50$} & \makecell{MRRE$_F$ \\ $k=10$} & \makecell{MRRE$_F$ \\ $k=50$} & \makecell{MRRE$_M$ \\ $k=10$} & \makecell{MRRE$_M$ \\ $k=50$} & Stead. & Cohev. & \makecell{KL Div. \\ $\sigma=1$} & \makecell{KL Div. \\ $\sigma=.1$} & \makecell{DTM \\ $\sigma=1$} & \makecell{DTM \\ $\sigma=.1$} & Stress\\
         \midrule
         UMAP           & \xtwo 0.9067 & \xtwo 0.8658 & \xone 0.9420 & \xsix 0.8773 & \xtwo 0.9113 & \xtwo 0.8922 & \xtwo 0.9524 & \xtwo 0.9227 & \xone 0.8538 & \xfour 0.6445 & \xfour 0.0042 & \xsix 0.2383 & \xsix 0.0662 & \xsix 0.4056 & \xfive 2.7369 \\
        UMAP (PCA) & \xone 0.9086 & \xone 0.8675 & \xtwo 0.9413 & \xfour 0.8885 & \xone 0.9137 & \xone 0.8943 & \xone 0.9526 & \xone 0.9267 & \xtwo 0.8491 & \xthree 0.6510 & \xthree 0.0034 & \xthree 0.2005 & \xfour 0.0579 & \xfive 0.3852 & \xsix 2.7735 \\
        UMAP (w/o ns)& \xsix 0.8471 & \xsix 0.8313 & \xthree 0.9294 & \xthree 0.8967 & \xsix 0.8496 & \xsix 0.8397 & \xfour 0.9347 & \xfour 0.9174 & \xsix 0.7439 & \xone 0.7469 & \xsix 0.0066 & \xfive 0.2120 & \xfive 0.0659 & \xfour 0.3808 & \xthree 0.9807 \\
        UMAP (w/o ns, PCA) & \xfour 0.8643 & \xfour 0.8416 & \xfour 0.9292 & \xone 0.8998 & \xfour 0.8724 & \xfour 0.8576 & \xthree 0.9394 & \xthree 0.9216 & \xfive 0.7706 & \xtwo 0.6977 & \xfive 0.0045 & \xfour 0.2069 & \xthree 0.0572 & \xthree 0.3684 & \xfour 0.9879 \\
         \textbf{UMATO} (Rand.) & \xfive 0.8619 & \xfive 0.8399 & \xsix 0.9180 & \xfive 0.8811 & \xfive 0.8650 & \xfive 0.8522 & \xsix 0.9231 & \xsix 0.9041 & \xthree 0.7805 & \xsix 0.5847 & \xtwo 0.0019 & \xtwo 0.1418 & \xtwo 0.0372 & \xtwo 0.3118 & \xone 0.8334 \\
         \textbf{UMATO} & \xthree 0.8716 & \xthree 0.8527 & \xfive 0.9266 & \xtwo 0.8989 & \xthree 0.8747 & \xthree 0.8627 & \xfive 0.9303 & \xfive 0.9150 & \xfour 0.7716 & \xfive 0.6178 & \xone 0.0015 & \xone 0.1290 & \xone 0.0348 & \xone 0.2915 & \xtwo 0.8391 \\
         \bottomrule
    \end{tabular}
    }
    \label{tab:accuvariants}
\end{table*}

\subsubsection{\rev{Comparison Against UMAP Variants}}

\label{sec:compvar}

\rev{

\boldsubsubsection{Objectives and design}
To identify which component of the UMATO algorithm contributes most to its competitive accuracy (\autoref{sec:compprac}), we focus on evaluating the impact of its two-phase optimization process (\autoref{sec:algorithm}).
The results of the previous experiment indicate the effectiveness of PCA initialization in improving the accuracy of UMATO. 

Here, we compare UMATO not only with the original UMAP but also with a variant that disables negative sampling, denoted UMAP (w/o ns). This variant of UMAP can also be interpreted as a form of UMATO in which all points are considered hub points and thus optimized without approximation. While this approach is impractical due to inefficiency, it could, in theory, show the optimal performance in preserving both global and local structure.
Comparing UMATO to this variant of UMAP thus reveals the contribution of UMATO's two-phase optimization design. For consistency, we use the same procedure and metrics as in the previous experiment (\autoref{sec:compprac}), with T\&C as the optimization target.

% We compare UMATO's accuracy 

% In the previous experiment, we find that UMATO's PCA initialization plays a crucial role in improving its accuracy in terms of preserving both local and global structure, but provide relatively small impact on UMAP's performance. 
% The results

\boldsubsubsection{Results and discussions}
The results are depicted in \autoref{tab:accuvariants}. UMAP (w/o ns) underperforms compared to the original UMAP in local structure preservation and to UMATO in global structure preservation. Contrary to our expectation, disabling negative sampling degrades the overall accuracy of UMAP. 
This degradation occurs because considering all pairwise distances between points during UMAP optimization introduces additional noise into the optimization process. The results clearly demonstrate the effectiveness of UMATO's two-phase optimization strategy.
}

\subsection{Scalability Analysis} 

\label{sec:scalexp}

We evaluate the scalability of UMATO. 
First, we compare all techniques using relatively small datasets. 
Then, we compare the top five scalable techniques with large datasets.
Finally, we investigate the runtime of individual stages of UMATO.

\subsubsection{Scalability Analysis with Small Datasets}

\label{sec:scalsmall}

\noindent
\textbf{Objectives and design.}
Our objective is to check whether UMATO can rapidly produce projections of small datasets. 
We apply UMATO and competitors to the 20 HD datasets we used in the accuracy analysis (\autoref{sec:accuexp}) and compare the runtime. 
To ensure robustness, we run each technique five times and record the average runtime. 
% We exclude LAMP from competitors as it costs $O(N^2)$, thus theoretically slower than other techniques.

\boldsubsubsection{Additional Competitor}
As UMATO's default initialization method (PCA) is substantially faster than the one used by UMAP (Spectral embedding), it may be unfair to compare these two algorithms directly with the default setting. We thus add UMAP with PCA initialization as an additional competitor.

\boldsubsubsection{Hyperparameter} 
For UMAP, LLE, PacMAP, Trimap, and UMATO, we set the number of nearest neighbors considered by the techniques to 15, which is the default value of UMAP. 
For UMATO, LAMP, and L-MDS, we set the number of hub points as 75, following the default of UMATO. For all other hyperparameters, we use the default value provided by the implementations. 

\boldsubsubsection{Apparatus}
We conduct the experiment using a Linux server equipped with Intel Xeon Silver 4210 and 224GB of RAM.

\boldsubsubsection{Results and discussions}
\autoref{fig:scal} depicts the results. 
While PCA shows the best scalability, UMATO is the runner-up, which is expected since UMATO incorporates PCA within its algorithm (\autoref{sec:algorithm}).
UMATO achieves $\times$ 4  performance improvement over UMAP regardless of the initialization method.
UMATO requires less than three seconds on average to generate projections. Such results validate UMATO's capability to promptly generate projections for small datasets, which will enhance its applicability in responsive and interactive systems.

\subsubsection{Scalability Analysis with Large Datasets}

\label{sec:scalsbig}

\begin{figure*}
    \centering
    \includegraphics[width=\linewidth]{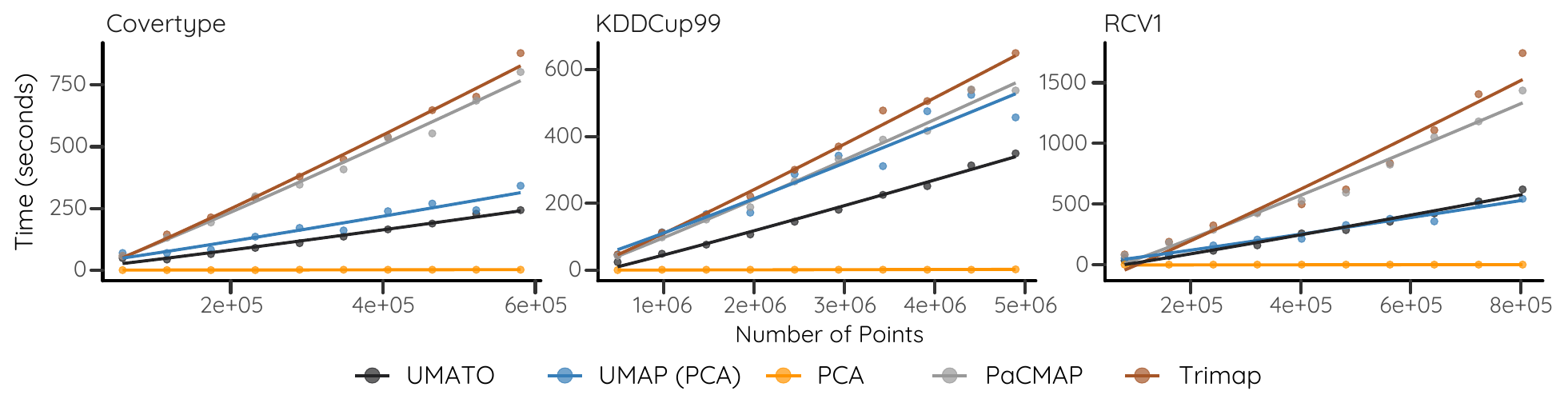}
    \vspace{-8mm}
    \caption{\rev{The results of the scalability analysis with large datasets (\autoref{sec:scalexp}). Overall, UMATO is on par with UMAP and outperforms every competitor except PCA. The regression line is fitted to the $y=a\cdot x\log x + b$ function, following the time complexity of UMATO, UMAP, and its variants (\autoref{sec:complexity}). LLE implementation is not depicted here as it requires more than 5,000 seconds to compute the smallest sampled subset of the data.}}
    
    \label{fig:scallarge}
\end{figure*}

% \begin{table}[t]
% \renewcommand{\arraystretch}{1.3}
%     \centering
%     \caption{The results of the scalability analysis with large datasets (\autoref{sec:scalexp}). 
%     The number of points (size) and dimensionality (dim.) are depicted on the right side of each dataset's name.
%     We depict the runtime of each DR technique in \texttt{mm:ss} format.
%     UMATO outperformed every competitor except PCA, with an average speedup of  $\times 14.3$ over UMAP.
%     Note that the LLE implementation failed to project all datasets due to the memory limit.
%     }
%     \scalebox{0.78}{
%     \begin{tabular}{lcccccc}
%     \toprule
%     Dataset (size $\times$ dim.) &   PCA & \makecell{UMAP\\ (Spectral)}  & \makecell{UMAP\\ (PCA)}& PacMAP & Trimap &\textbf{UMATO} \\
%     \midrule
%     Covertype  (581K $\times$ 54) & 00:01 & 71:11 & 03:55 & 06:48 & 07:27 &  04:08\\
%     RCV1       (804K$\times$ 50) & 00:01& 87:22 & 06:38 & 18:10 & 21:49 & 08:56\\
%     KDDCup99   (4.90M $\times$ 41) & 00:01 & 31:36 & 04:30 & 07:27 & 09:40 & 04:59 \\
%     \bottomrule
%     \end{tabular}
%     }
%     \label{tab:scallarge}
% \end{table}

\noindent  
\textbf{Objectives and design.}
We aim to further verify UMATO's scalability by testing it on large datasets. We prepare three datasets with more than 500K data points and check the time needed for UMATO and competitors to process the datasets.
To ensure the experiment ends in a reasonable time,
we use UMATO and alternative DR techniques that ranked in the top five scalabilities in the previous experiment with small datasets (UMAP with PCA initialization, LLE, PCA, PacMAP, Trimap; refer to \autoref{sec:scalsmall}). 
% We also added UMAP with Spectral embedding initialization as a competitor, as it is the default setting of UMAP. 
\rev{
We test these competitors on the original datasets and their subsampled versions to examine how runtime varies with sample size. We adjust the sampling rate from 10\% to 100\% in 10\% increments, with each technique executed once per sampled dataset.
}
We use the same hyperparameter and apparatus setting as in the previous experiment (\autoref{sec:scalsmall}).

\boldsubsubsection{Datasets}
We use Covertype \cite{asuncion07uci}, KDDCup99 \cite{asuncion07uci}, and RCV1 \cite{lewis2004rcv1} datasets. For RCV1, we reduce the dimensionality from 47K to 50 because the original dataset is represented in a compressed sparse row format, which is incompatible with PacMAP and UMATO implementations. 

\boldsubsubsection{Results and Discussions} As shown in \autoref{fig:scallarge}, UMATO \rev{is the runner-up after PCA.}
\rev{It performs comparably to UMAP with PCA initialization and outperforms PaCMAP and Trimap in scalability.}
The fact that UMATO outperforms Trimap and PaCMAP, two scalable variants of UMAP that also use PCA for initialization, strongly supports UMATO's advantage in terms of scalability. \rev{This trend remains consistent across varying dataset sizes.} 

% places in third place, slightly slower than UMAP with PCA initialization. 
% UMATO, on the other hand, achieves a substantial speedup compared to UMAP with Spectral embedding initialization. 
% The results verify that the UMATO is on par with or slightly slower than UMAP, which aligns with our computational complexity analysis (\autoref{sec:complexity}).
% Furthermore, the fact that UMATO outperforms Trimap and PacMAP, two scalable variants of UMAP that also use PCA for initialization, strongly verifies UMATO's advantage in terms of scalability.

% These results validate UMATO's scalability in processing large datasets, reaching millions of points. 

\begin{figure}
    \centering
    \includegraphics[width=\linewidth]{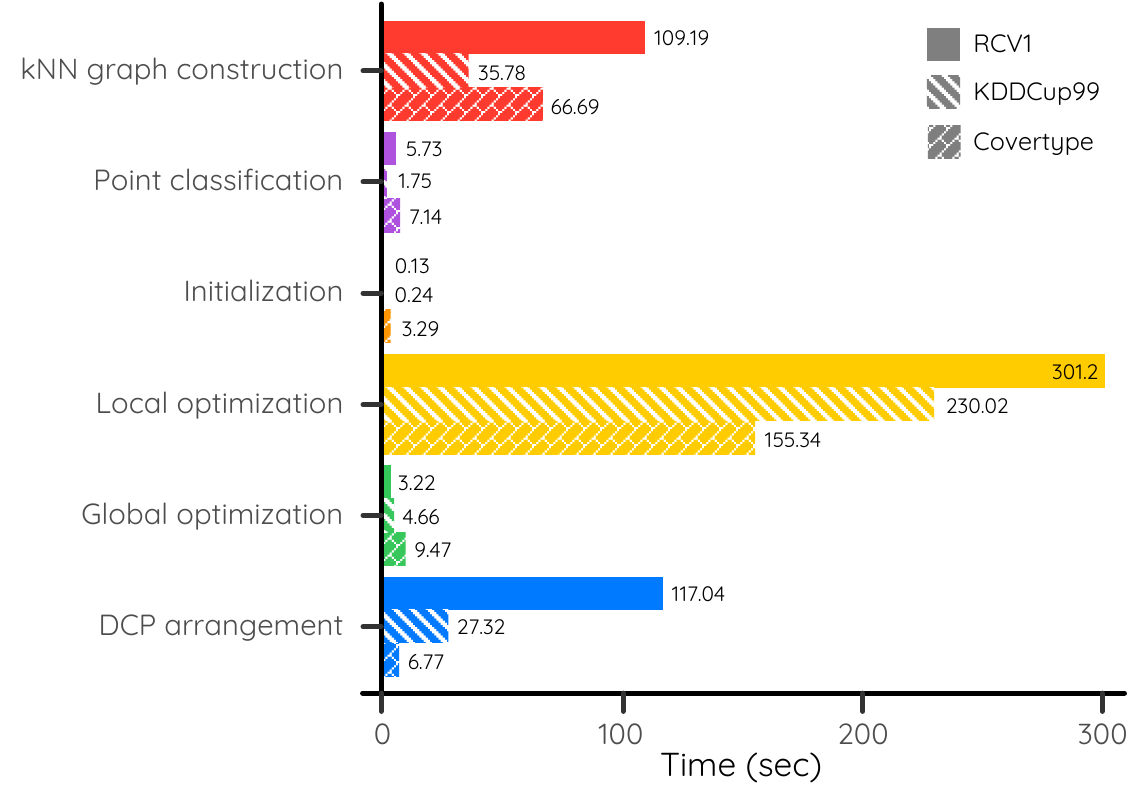}
        \vspace{-6mm}
    \caption{Runtime of individual stages in UMATO. Overall, local optimization and $k$NN graph construction dominate the runtime. In terms of RCV1 dataset, DCP arrangement also requires substantial time.}
    \label{fig:indrun}
\end{figure}

\subsubsection{Scalability Analysis for Individual Stages}

\label{sec:inidividualruntime}

\boldsubsubsection{Objectives and design}
We investigate the runtime of individual stages of UMATO, thereby identifying the bottleneck of the technique. We check the runtime required to compute each stage of UMATO (\autoref{sec:algorithm}): $k$NN graph construction, point classification, initialization, global optimization, local optimization, and DCP arrangement. We use the same hyperparameters, datasets, and apparatus as in previous experiments (\autoref{sec:scalsmall}, \ref{sec:scalsbig}).

\boldsubsubsection{Results and discussions}
The results are depicted in \autoref{fig:indrun}. For all three datasets, we identify that $k$NN graph construction and local optimization stages dominate the runtime of UMATO. This result reaffirms our computational complexity analysis (\autoref{sec:complexity}), where these two stages theoretically dominate the computation of UMATO. 
Further optimizing these stages will be essential to enhance the usability of UMATO. We will discuss possible directions in \autoref{sec:limitations}.

We also find that the ratio of DCP arrangement among total runtime is notably higher in the RCV1 dataset than in the other two datasets. The result indicates that UMATO may require larger computation for outlier-rich datasets. Combining UMATO with outlier detection and removal algorithms \cite{boukerche20cs} to reduce runtime will be an interesting future avenue to explore.

\subsection{Stability Analysis} 

\label{sec:stabexp}

We evaluate the stability \cite{jung2025arxiv} of UMATO and baseline techniques against two common data perturbations in DR: subsampling and initialization. Here, we hypothesize that UMATO will exhibit high stability, supporting more reliable data analysis. This is because the global optimization step of UMATO, which determines the overall shape of the resulting projection, runs without any approximations (\autoref{sec:layoutopt}).

\subsubsection{Stability Against Subsampling}

\label{sec:stabsubsample}

\boldsubsubsection{Objectives and design}
We aim to evaluate the stability of UMATO against data subsampling. 
Subsampling is a common strategy for obtaining DR results in a reasonable time by sampling a portion of the original dataset and running DR on the subsample. 
The primary concern is whether a subsampled projection can accurately represent the patterns in the original dataset. For subsampling to be reliably used in practice, the projection of a subsampled dataset should be comparable to the subsample of the projection made from the original dataset.

The stability against data subsampling is measured by evaluating the geometric similarity between the projection of a subsample and the subsample of the projected points made with the entire dataset. 
We use a Procrustes analysis for this purpose.
First, we align two projections by applying a permutation that best aligns them. We apply permutation first since the two projections being compared can consist of different points in the original space. 
Then, translation, uniform scaling, and rotation are applied to the two projections. 
Finally, we compute the Procrustes distance between two projections.
For two projections $X = \{x_1, x_2, ..., x_n\}$ and $Y = \{y_1, y_2, ..., y_n\}$, Procrustes distance is defined as:
\begin{equation}    \label{eq:procrustes}
    d_P(X, Y) = \sqrt{\sum_{i = 1}^{n} (x_i - y_i) ^ 2}.
\end{equation}
A distance of 0 indicates a perfect match, while a distance of 1 indicates maximum dissimilarity. 

To comprehensively evaluate the stability of DR techniques, we conduct Procrustes analysis on diverse datasets and sampling rates. For each pair of datasets and DR technique, we conduct the analysis 50 times, where the sampling rate is randomly selected between 10\% and 99\%.

\boldsubsubsection{Hyperparameter} We use the same hyperparameter setting as in the scalability analysis (\autoref{sec:scalexp}).

\boldsubsubsection{Datasets}
Among the 20 collected datasets, we exclude those with sizes smaller than $3,000$, as they do not result in sufficient subsample sizes. As a result, we use seven datasets in total. The datasets used in this experiment are underlined in \autoref{tab:datasets}.

\begin{figure}[t]
    \centering
    \includegraphics[width=\linewidth]{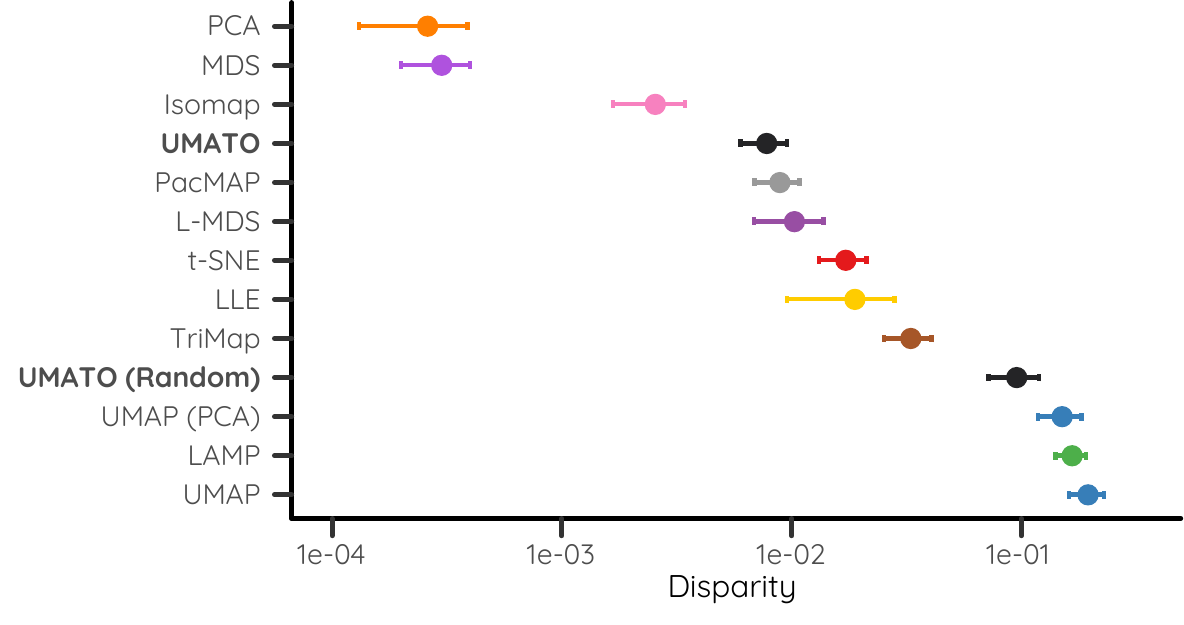}
    \vspace{-8mm}
    \caption{The stability of UMATO and baseline techniques against subsampling (\autoref{sec:stabsubsample}). The smaller the disparity is, the more stable the corresponding DR technique is. 
    Error bars depict 95\% confidence intervals. }
    \label{fig:stab_subsample}
    % \vspace{-5mm}
\end{figure}

\begin{figure}[t]
    \centering
    \includegraphics[width=\linewidth]{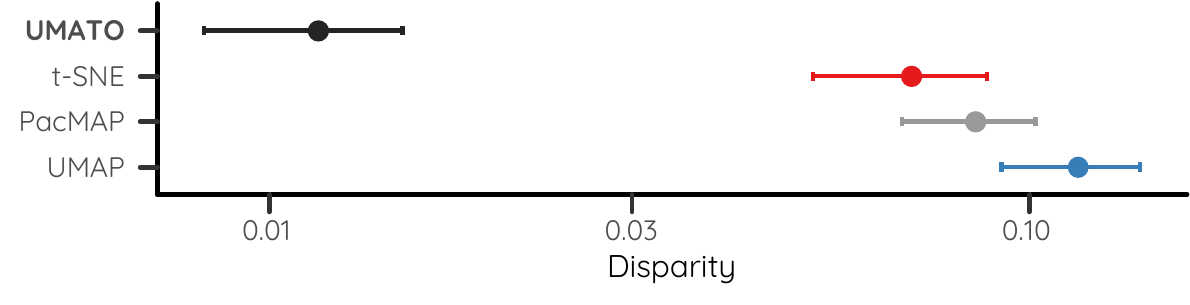}
    \vspace{-8mm}
    \caption{The stability of UMATO and competitors against diverse initialization method (\autoref{sec:stabinitial}). The smaller the disparity is, the more stable the corresponding DR technique is. 
    Error bars depict 95\% confidence intervals.
    Among the four DR techniques we compare, UMATO showed the best stability over the change of initialization method. }
    \label{fig:stab_init}
    % \vspace{-5mm}
\end{figure}

\begin{figure*}[h]
    \centering
    \includegraphics[width=\textwidth]{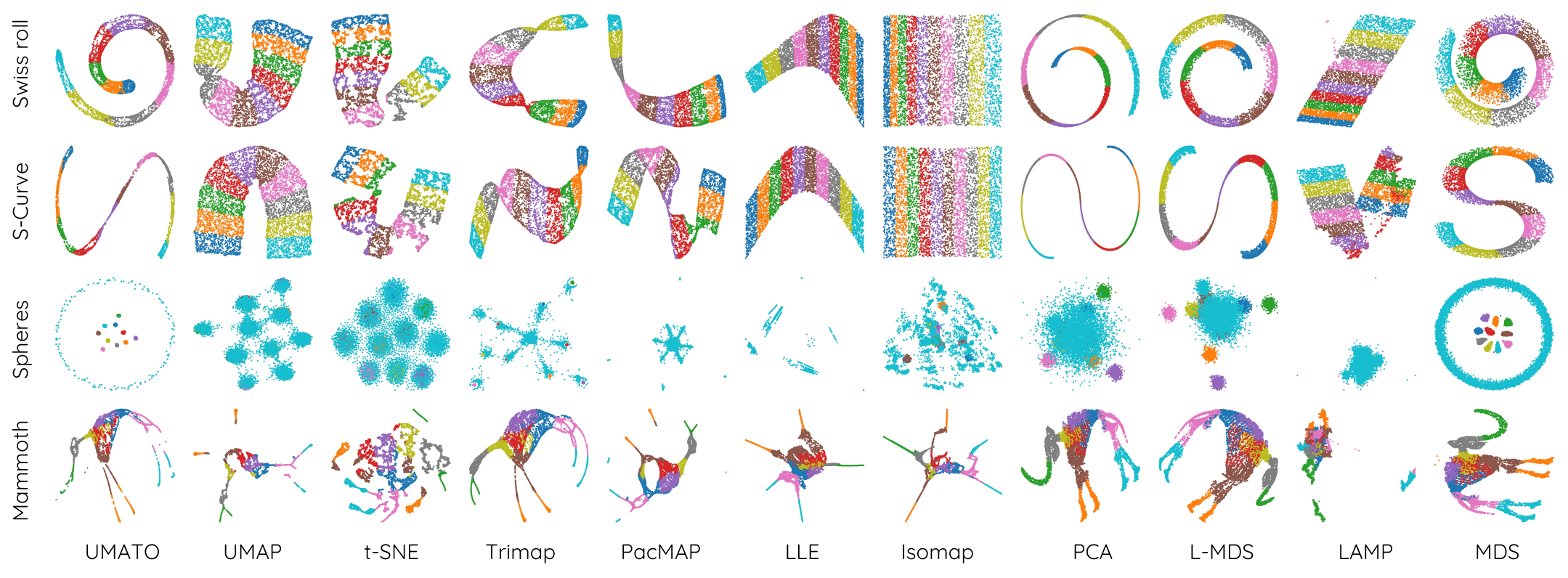}
    \vspace{-7mm}
    \caption{The projections used in our qualitative experiment (\autoref{sec:qualexp}). While UMATO succeeds in accurately depicting the original structure for all four datasets, competition techniques fail to do so.}
    \label{fig:syn_embeddings}
\end{figure*}

\boldsubsubsection{Results and discussions}
\autoref{fig:stab_subsample} presents the results. UMATO shows the best stability against subsampling except for PCA and Isomap. Moreover, UMATO is up to ten times more stable than UMAP. 
PCA, Isomap, \rev{and MDS} outperform UMATO as they are techniques that rely on matrix multiplication in reducing dimension.
Since these transformation matrix depends on the data features, they are inherently robust to data subsampling, much like data variance. 
Nonetheless, the fact that UMATO outperforms all other nonlinear techniques and even a linear technique (L-MDS) validates the effectiveness of our two-phase optimization scheme and the reliability of visual analytics using UMATO.

\rev{The results also imply the importance of PCA initialization in improving stability. We find that UMATO is ten times less stable with random initialization. However, UMAP shows negligible improvement due to PCA initialization, which aligns with the results of our accuracy analysis (\autoref{sec:compprac}). The results clearly indicate the positive interplay between PCA initialization and the two-phase optimization scheme of UMATO.}

% linear DR techniques (PCA, L-MDS) generally outperform nonlinear DR techniques. This is because linear techniques rely on matrix multiplication in reducing dimension, where the transformation matrix depends on the data features robust to data subsampling, e.g., data variance. Still, the fact that UMATO outperforms all other nonlinear techniques and L-MDS validates the effectiveness of our two-phase optimization scheme. 

\subsubsection{Stability Against Initialization Method}

\label{sec:stabinitial}

\noindent
\textbf{Objectives and design.}
We aim to evaluate the projection stability of UMATO against initialization methods. 
It is widely known that the characteristics of DR projections highly depend on initialization \cite{kobak2021initialization}. 
The more sensitive a DR technique is to initialization, the less reproducible the data analysis based on that technique becomes. Therefore, robustly producing stable projections regardless of the initialization methods is essential for a reliable DR technique.

The evaluation process is as follows:
For a given dataset and a DR technique, we generate five projections with diverse initialization methods. Three are randomly initialized, and the remaining two are initialized using PCA and Spectral embedding. We select PCA and Spectral embedding because they are the default initialization methods for UMATO and UMAP, respectively. 
We then perform Procrustes analysis (see \autoref{sec:stabsubsample} for details) on the projections in a pairwise manner, using the resulting scores as a proxy for the stability of the corresponding DR technique. 

\boldsubsubsection{Competitors}
We compare nonlinear DR techniques that have an initialization process followed by an optimization step.
We also exclude the competitors whose implementations do not allow changes to the initial projection. 
As a result, we compare UMATO against UMAP, $t$-SNE, and PacMAP.

\boldsubsubsection{Hyperparameter and datasets}
We use the same hyperparameter setting and datasets as in the scalability analysis (\autoref{sec:scalexp}).

\boldsubsubsection{Results and discussions} 
\autoref{fig:stab_init} depicts the results. Among four DR techniques that share the initialization and following optimization process, UMATO shows the best stability. Compared to UMAP, UMATO is up to 10 times more stable. 
As in the stability analysis over subsampling (\autoref{sec:stabsubsample}), these results clearly verify that using UMATO will substantially enhance the reliability of HD data analysis.

%% file: sections/06_demonstration.tex
\begin{figure*}[t]
    \centering
    \includegraphics[width=\linewidth]{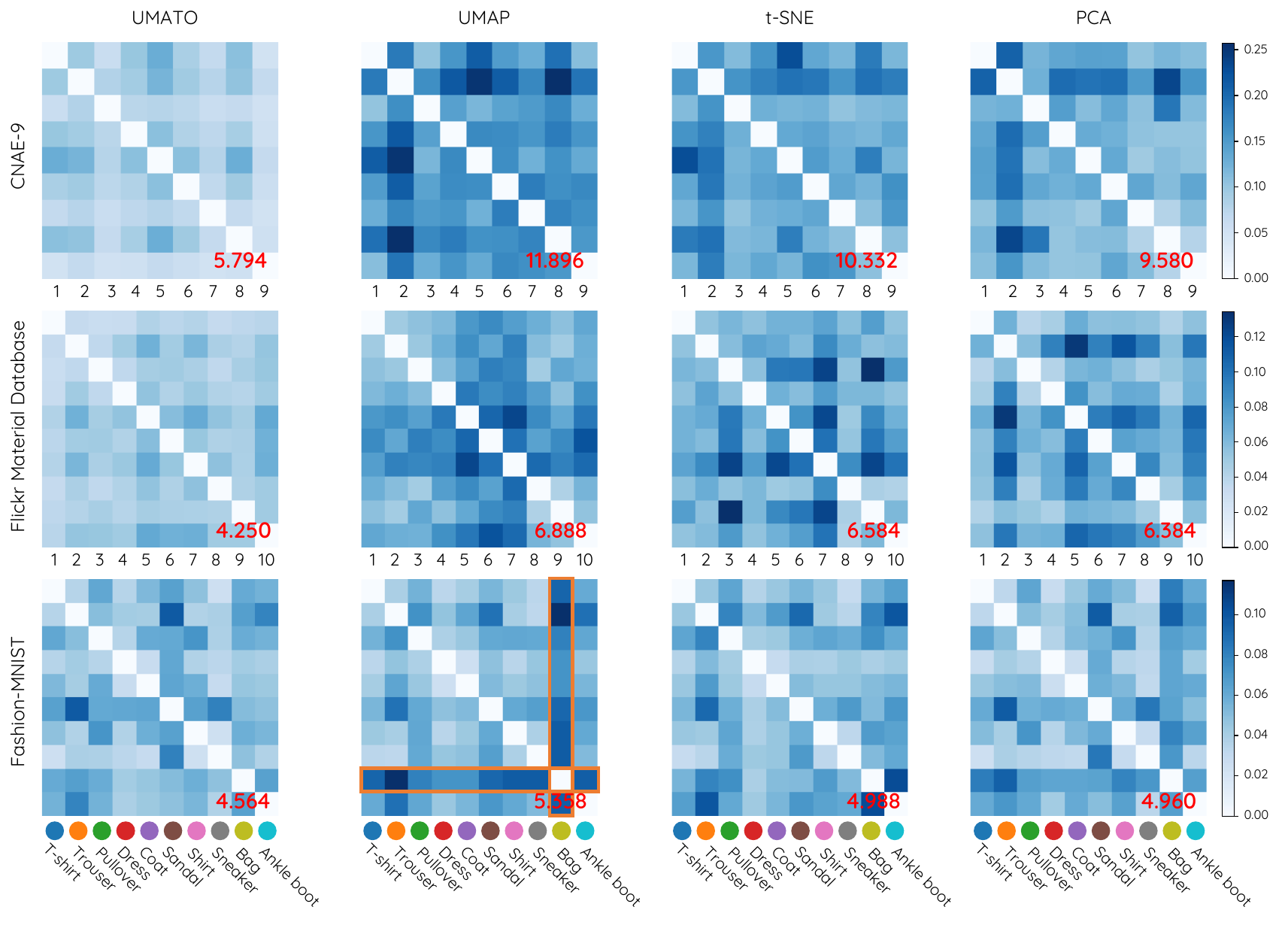}
    \vspace{-7mm}
    \caption{
    Heatmaps representing how well the relationships between each pair of classes are preserved by four DR techniques (UMATO, UMAP, $t$-SNE, and PCA) (\autoref{sec:usecase}). Each cell depicts the KL divergence score locally computed for the corresponding pair of classes (the lower, the brighter and better). The colors are normalized across each dataset (row).
    The red numbers depicted in the lower right corner of each heatmap represent the sum of scores across the heatmap. Overall, UMATO performs best in preserving pairwise relationships between classes, indicating its effectiveness in supporting reliable analysis of labeled data. 
    }
    \label{fig:usecase_heatmap}
\end{figure*}

\begin{figure}
    \centering
    \includegraphics[width=\linewidth]{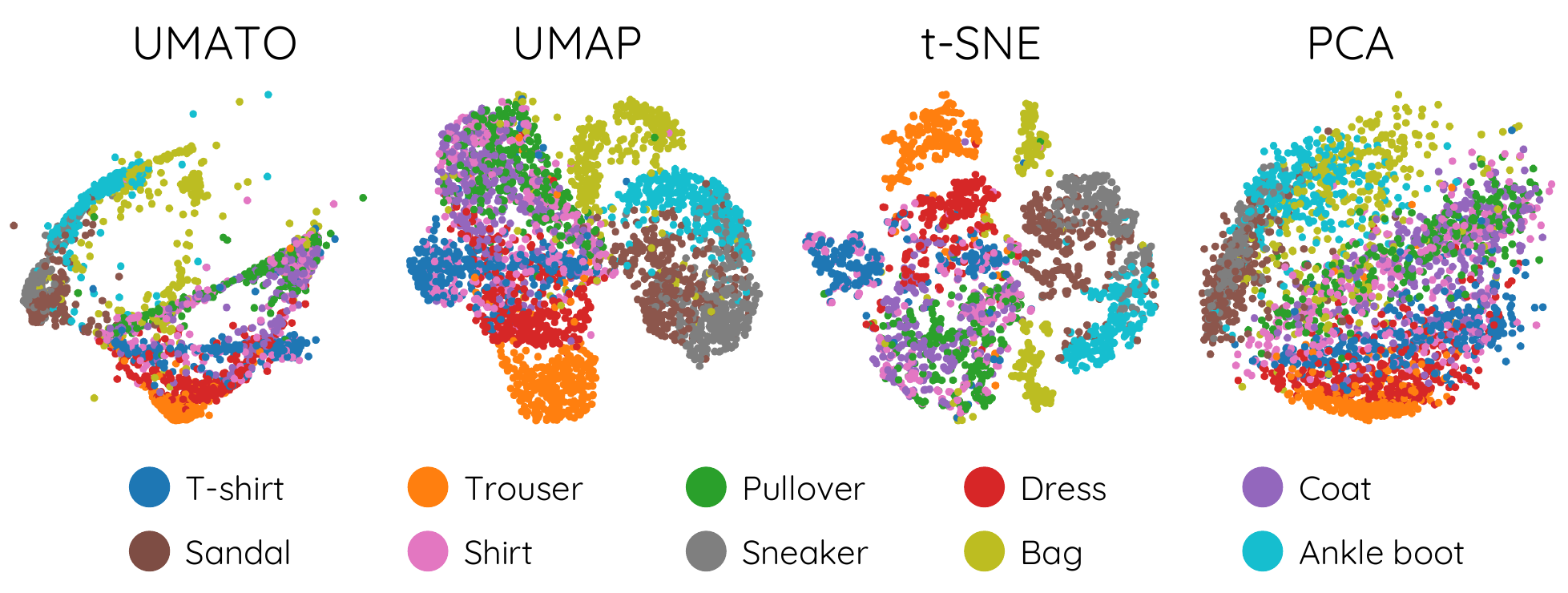}
    \vspace{-7mm}
    \caption{
    UMATO, UMAP, $t$-SNE, and PCA projections of Fashion-MNIST dataset. Our case study (\autoref{sec:usecase}) demonstrates that UMATO projections can help analysts analyze the global relationship between class labels in a reliable manner.  
    }
    \label{fig:usecase_projs}
\end{figure}

\begin{figure*}
    \centering
    \includegraphics[width=\linewidth]{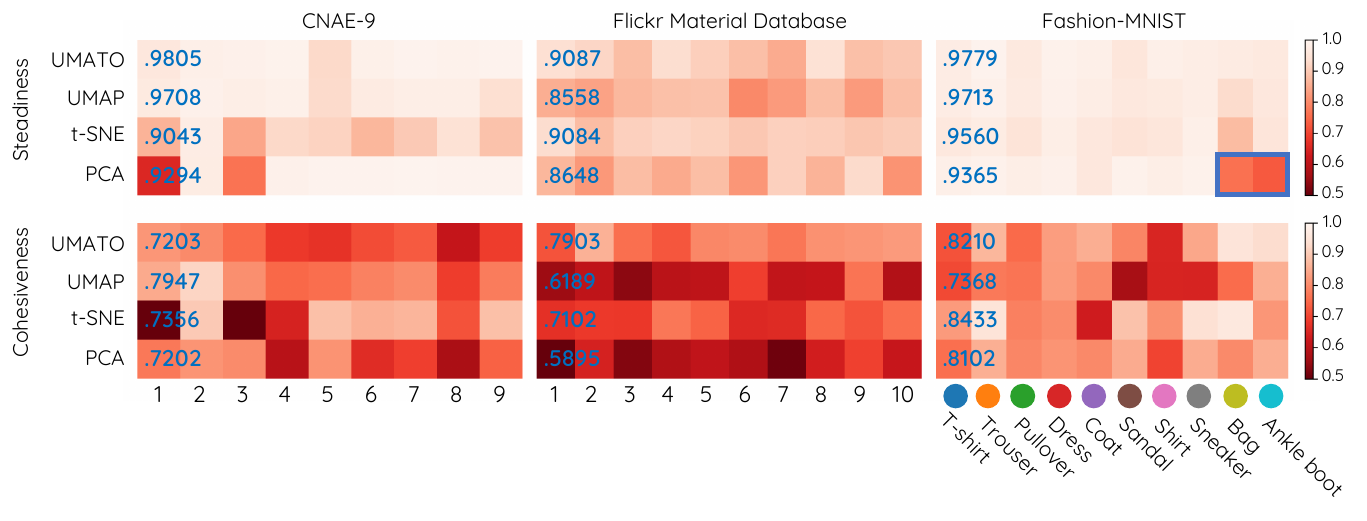}
    \vspace{-9mm}
    \caption{
    \rev{Heatmaps representing how well the structure of each class is maintained by the projections generated by UMATO, UMAP, $t$-SNE, and UMAP. Each cell depicts the average Steadiness \& Cohesiveness (S\&C) score of the points within each class (the higher, the brighter and better), and the blue number in each row shows the average S\&C score across the classes.}
    }
    \label{fig:usecase_snc}
\end{figure*}

\section{Demonstration}

\label{sec:qualexp}

We qualitatively verify that by focusing both on global and local structures, UMATO faithfully represents the manifold structure of HD data.
To do this, we prepare diverse synthetic datasets with known structures. 
We then apply UMATO and baseline techniques (UMAP, $t$-SNE, Trimap, PCA, PacMAP, LLE, L-MDS, LAMP) and manually investigate whether the projections accurately depict the original characteristics of the data.
Following our accuracy analysis (\autoref{sec:accuexp}), we use Bayesian optimization \cite{snoek12nips} with T\&C loss function to generate optimal projections. 

\subsection{Datasets}
We utilize four synthetic datasets.
The brief description of each dataset and the salient structural characteristics that any effective DR techniques should preserve are as follows:

\boldsubsubsection{Swiss roll} 
This dataset consists of a plane rolled into the 3D space. We generate the Swiss roll consisting of 5,000 points using \texttt{scikit-learn} library. An effective DR technique may accurately represent both the structure of the plane and its global structure (i.e., rolled shape). 

\boldsubsubsection{S-Curve} The dataset is similar to the Swiss roll, but the plane is curved into an S-shape instead of a roll. Like the Swiss roll, we made an S-curve with 5,000 points with \texttt{scikit-learn} library. An effective DR technique should accurately represent both the plane and its global curved shape. 

\boldsubsubsection{Mammoth}
The Mammoth dataset \cite{Coenen2019} is a 3D point cloud representing the skeleton of a mammoth. Among the different versions provided by Coenen and Pearce \cite{Coenen2019}, we use the one consisting of 10,000 points. We aim to check whether DR projections accurately depict the real appearance of the mammoth. 

\boldsubsubsection{Spheres}
This dataset, first introduced by Moor et al. \cite{Moor19Topological}, consists of 101-dimensional spheres. Ten small spheres, each with a radius containing 500 points, are enclosed by a large sphere with 5,000 points. We expect an effective DR projection to accurately reflect the inclusion relationship between the small and large spheres.
% The original structure of Swiss rolee, S-Curve, and Mammoth datasets are visualized in \autoref{fig:syn_dataset}.
We do not depict the Spheres dataset as it lies in the 101-dimensional space.

\subsection{Qualitative Analysis}

The resulting projections are depicted in \autoref{fig:syn_embeddings}. 
For Swiss roll and S-curve datasets, UMATO, L-MDS, and MDS capture the global structure(rolled and curved shapes) while unrolling the local plane structure.
UMAP, $t$-SNE, Trimap, PacMAP, LLE, Isomap, and LAMP accurately depict the dataset as planes (capturing the local structure) but fail to capture the global shapes.
In contrast, PCA successfully captures the global structure but often represents local manifolds as lines instead of planes.

In terms of the Mammoth dataset, UMATO, PCA, PacMAP, L-MDS, and MDS succeed in accurately representing the overall characteristics of the Mammoth skeleton. $t$-SNE  and LAMP totally lose the structure. UMAP, Trimap, Isomap, and LLE preserve local structures, but their global arrangement is distorted.  

For the Spheres dataset, UMATO and MDS accurately represent the relationship between the outer and inner spheres. 
In their projections, we can find that the outer circle encloses inner spheres in a circular form, providing an intuitive depiction of the original global structure. 
In contrast, other baseline techniques failed to accurately depict the inclusion relationship. 
For example, in the UMAP projection, a big enclosing hypersphere is divided and merged into small hyperspheres. This occurs because UMAP focuses on local neighborhood structure and thus hardly captures the existence of a big hypersphere. A similar phenomenon occurs in PacMAP, t-SNE, Isomap, and Trimap. 
In contrast, in PCA and L-MDS projections, the inner spheres are located outside the outer sphere, which is a totally incorrect representation of the original dataset.

In summary, UMATO faithfully represents the overall manifold structure of all four datasets. 
This qualitatively reaffirms the results of our accuracy analysis (\autoref{sec:accuexp}), demonstrating UMATO's superiority in reliable visual analytics of HD data.

%% file: sections/07_use_case.tex
\section{Case Study}

\label{sec:usecase}

We present a case study with real-world datasets demonstrating how UMATO contributes to the reliable analysis of HD data. 

\subsection{Objectives and Design}

We showcase the effectiveness of UMATO in supporting reliable analysis of labeled datasets. We simulate a situation in which an analyst generates DR projections of a given labeled dataset and visualizes them using scatterplots, where the color of each point depicts the corresponding class label. 
We assume that the analyst wants to investigate the relationship between class labels, e.g., overlap or separation between classes \cite{etemadpour15tvcg, xia2021revisiting}, which is a common task in labeled scatterplots \cite{lu20tvcg, jeon24tvcg, yunhai18tvcg}.
We project datasets using DR techniques, including UMATO, and then quantitatively examine how well pairwise relationships between class labels are preserved. The detailed setting we use is as follows:

\boldsubsubsection{DR projections}
We compare four DR techniques: UMATO, UMAP, $t$-SNE, and PCA. We select UMAP, $t$-SNE, and PCA as competitors as they are widely used DR techniques nowadays \cite{xia2021revisiting} and also show the top or runner-up performance in preserving local or global structures in our accuracy analysis (\autoref{sec:accuexp}, \autoref{fig:rank_accuracy}). 
To guarantee fair comparison across DR techniques, we optimize the hyperparameters of the techniques using Bayesian optimization. 
Considering the assumption that the analyst wants to investigate the relationship between class labels, we use a global metric (KL divergence with $\sigma=0.1$) as an optimization target. 

\boldsubsubsection{Evaluating the preservation of the relationship between classes}
\rev{We want to evaluate whether the projections reliably support the target task---which is to investigate the relationship between classes. We achieve this by evaluating whether the separability between classes is maintained. To do so, we apply}
KL divergence for each pair of classes. Formally, for a given HD dataset $X = \{x_1, x_2, \cdots, x_n\}$ and a corresponding projection $Y = \{x_1, x_2, \cdots, x_n\}$, we construct a matrix $M$ where $(i,j)$-th cell $M_{i, j}$ \rev{is defined as}: 
\[
M_{i, j} = \left\{
\begin{array}{rcl}
0 & \text{if} & i = j \\
KL (C(X, \{i, j\}), C(Y, \{i, j\})) & \text{if} & i \neq j
\end{array}
\right..
\]
Here, $C(Z, \{i, j\})$ represents the subset of data $Z$ having label $i$ or $j$ and $KL$ represents KL divergence. Note that lower values in matrices indicate better performance of $Y$ in preserving the relationship between classes. 

\rev{However, KL divergence only explains whether the separability between classes in the HD space are well represented in the projection or not. For more comprehensive analysis, we use S\&C, quality measures specifically designed to examine overlap and separation between clusters \cite{jeon2022measuring} (\autoref{sec:accuexp}). To examine how the representation of each class is distorted, we first compute the degree to which each point is distorted using S\&C, then aggregate these scores in a class-wise manner. A low Steadiness score means that the classes overlap with other classes or their density is overrepresented. Conversely, low Cohesiveness means that the separability between classes is exaggerated or their density is underrepresented \cite{jeon2022measuring}.}

\boldsubsubsection{Datasets}
We prepare three datasets: CNAE-9 \cite{asuncion07uci}, Flicker Material Database \cite{sharan2009material}, and Fashion-MNIST \cite{xiao2017-online}. We use these datasets because they have a sufficient number of class labels (nine, nine, and ten for each), making them suitable for simulating our assumed situation.

\subsection{Result and Discussions}

\autoref{fig:usecase_heatmap} depicts the heatmaps representing $M$s computed across four DR techniques and three datasets. Overall, UMATO shows the best performance (lighter color) in preserving the relationship between pairs of class labels for all three datasets. The outcome indicates that UMATO projections help analysts the most in reliably examining class relationships.

The results verify the effectiveness of balancing global and local structures in achieving reliable visual analytics using DR. As seen in \autoref{fig:usecase_projs}, UMAP and $t$-SNE well separate class labels. This is because these techniques focus on local structure, thus exaggerating the distance between non-neighboring points \cite{lee2007nonlinear, jeon24tvcg}.
\rev{However, a close examination of KL divergence and S\&C scores suggests} that this separation may be misleading. For example, in UMAP's \rev{projection of the Fashion-MNIST dataset} (\autoref{fig:usecase_projs}), the \textit{Bag} class is placed distinctly from other classes. 
Still, the corresponding heatmap \rev{representing KL divergence scores} shows that the relationships between \textit{Bag} and other classes are inaccurately presented (solid orange boxes in \autoref{fig:usecase_heatmap}), indicating that such a distinction can be misleading. 
This result also aligns with the findings from previous literature \cite{wattenberg2016how, Coenen2019}. 
\rev{In the PCA projection, we observe greater overlap between class labels (\autoref{fig:usecase_projs}); however, the poor KL divergence score indicates that such overlap is misleading. S\&C scores reveal that this distortion primarily originates from the \textit{Bag} and \textit{Ankle boot} classes (\autoref{fig:usecase_snc}, solid blue box). Note that the same pattern---a substantial contribution of specific classes to overall distortions---is also observed in the CNAE-9 dataset (\autoref{fig:usecase_snc} first row, first column).}
This is because PCA cares less about local structures; thereby, non-neighboring points are likely to be projected in similar locations \cite{jeon2022measuring, nonato2018multidimensional}.
Conversely, UMATO projection shows an intermediate level of class overlap and achieves high average scores in both KL divergence and S\&C, showing the best performance in preserving class-pairwise relationships overall.
We can attribute this outcome to UMATO's preservation of global arrangements between classes while avoiding false neighbors by considering local structures.

%% file: sections/08_hyperparameter.tex
\begin{figure*}
    \centering
    \includegraphics[width=\linewidth]{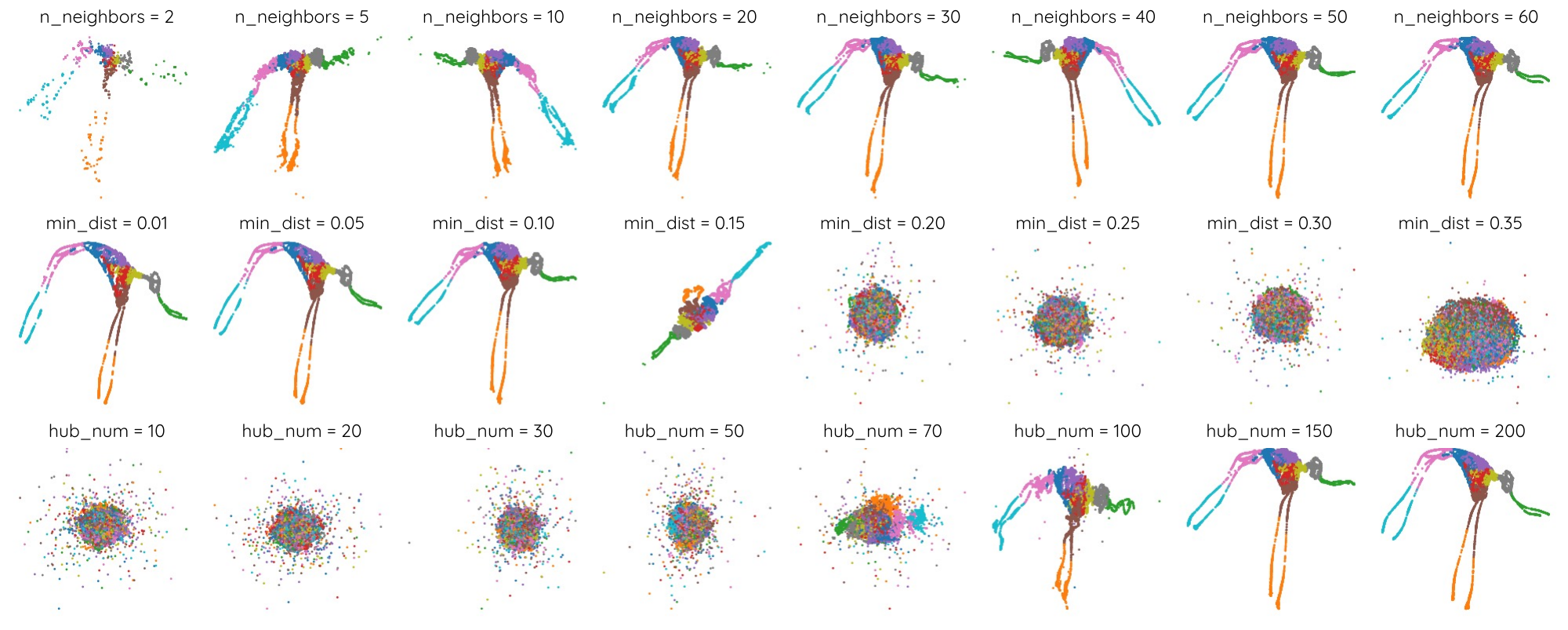}
    \vspace{-7mm}
    \caption{Illustration of how three major hyperparameters (\texttt{min\_dist}, \texttt{n\_neighbors}, \texttt{hub\_num}) in UMATO affect the projections of the Mammoth dataset. 
    The projections are made by tweaking a single hyperparameter value from the default hyperparameter setting \texttt{min\_dist}: 0.1, \texttt{n\_neighbors}: 50, \texttt{hub\_num}: 150). The value of the tweaked hyperparameter is depicted above each projection.
    To produce reliable projections, we need to use a small \texttt{min\_dist} (second row) and a sufficiently high \texttt{hub\_num}. If these conditions are met, UMATO produces projections with similar structures regardless of hyperparameter values.}
    \label{fig:hp}
\end{figure*}

\section{Effects of Hyperparameters in UMATO}

\label{sec:hp}

We describe how the hyperparameters of UMATO affect the resulting projections as a qualitative guideline to set hyperparameters in practice.
We focus on \texttt{n\_neighbors} and \texttt{min\_dist}, the hyperparameters originating from UMAP.
While \texttt{n\_neighbors} denotes the number of NN considered in the graph construction step (\autoref{sec:knnconst}), \textit{min\_dist} denotes the minimum distance between data points in the projection.
We also focus on \texttt{hub\_num}, a hyperparameter representing the number of hubs considered in global layout optimization (\autoref{sec:layoutopt}). We empirically find that other hyperparameters' (e.g., $a$ and $b$ in \autoref{eq:pos}) effect is negligible compared to these three hyperparameters.

\boldsubsubsection{\texttt{n\_neighbors}}
It is widely known that \texttt{n\_neighbors} determines the degree to which UMAP focuses on global structure \cite{Coenen2019, mcinnes2018umap}. While low \texttt{n\_neighbors} makes UMAP more focused on the fine-grained local structure, high value makes it better represent the global structure. We find that UMATO also focuses more on local structure when \texttt{n\_neighbors} is small. 
For example, in the first row of \autoref{fig:hp}, low \texttt{n\_neighbors} leads to projections with relatively small clusters. This is because an insufficient number of \texttt{n\_neighbors} makes the algorithm interpret local clusters as a set of loosely connected components. Such a phenomenon also occurs in UMAP \cite{Coenen2019}. 

However, even with low \texttt{n\_neighbors}, the global structure of HD data is well preserved. As seen in \autoref{fig:hp}, regardless of \texttt{n\_neighbors} value, UMATO preserves the global shape of the Mammoth skeleton. Such results validate the effectiveness of our two-phase optimization scheme in preserving global structure robustly.

\boldsubsubsection{\texttt{min\_dist}}
In UMAP, this hyperparameter controls the clumpiness of projections; smaller \texttt{min\_dist} values lead to tightly condensed clusters. 
In contrast, such an effect is minimized in UMATO. For example, in the second row of \autoref{fig:hp}, small \texttt{min\_dist} does not change the overall compactness of clusters. This is because the global optimization (\autoref{sec:layoutopt}), which determines the overall shape of the projection, is executed with a relatively small number of points. 
Regardless of the decrease in \texttt{min\_dist}, the pairwise distances of these points are sufficiently larger than \texttt{min\_dist}, and therefore, the global structure of the projection remains unchanged.

However, we find that the overall structure of projections suddenly collapses when \texttt{min\_dist} increases beyond a certain threshold (\autoref{fig:hp}, second row). We investigate that not only local structures but also global structures are distorted in these projections, implying that high \texttt{min\_dist} values disturb global optimization. 
This is because a high \texttt{min\_dist} value makes hub points uniformly distributed across the projection space, thereby marginally capturing the true global structure of the original HD data.
In conclusion, we recommend using a low \texttt{min\_dist} value to produce reliable projections in practice. We empirically find that the value around 0.1 produces reliable projections overall.

\begin{figure}
    \centering
    \includegraphics[width=\linewidth]{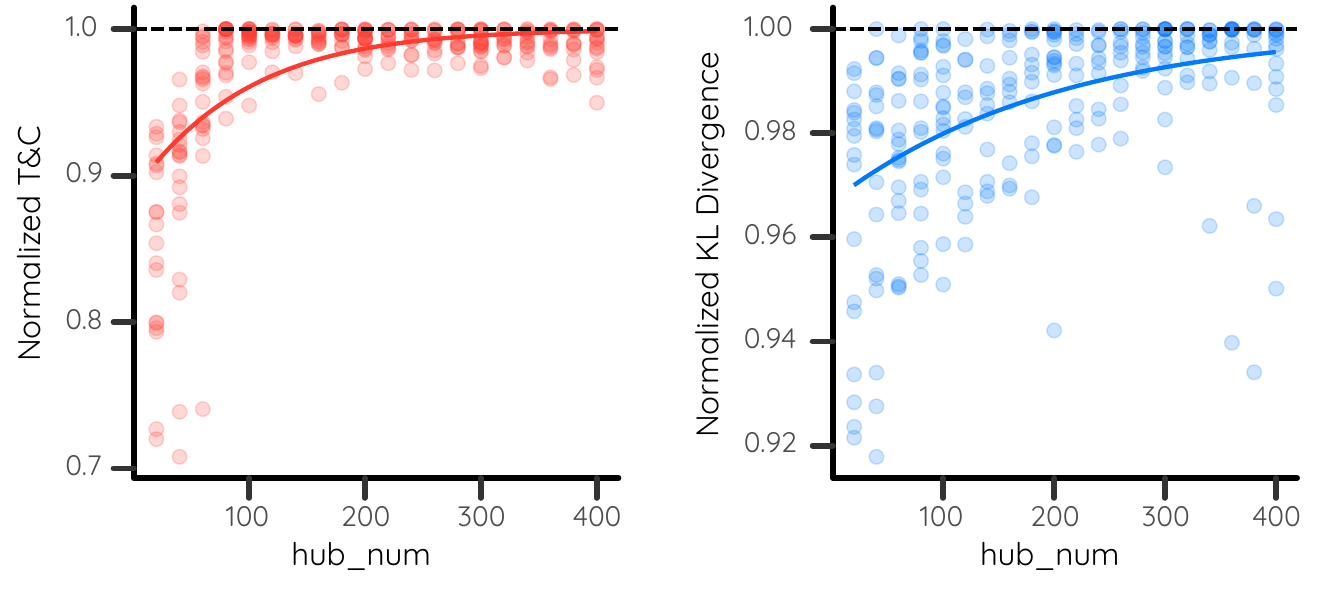}
    \vspace{-7mm}
    \caption{
       Normalized T\&C and KL divergence scores of UMATO projections with different \texttt{hub\_num}. The scores are normalized by dividing each score by the maximum score obtained within its respective dataset. Note that we use the value subtracted from 1 for KL divergence to max bigger values to indicate better projections. 
       Trend lines are fitted following the logistic function. While T\&C scores converge to the maximum value around \texttt{hub\_num} $=200$, KL divergence scores do not converge until \texttt{hub\_num} exceeds 350. 
    }
    \label{fig:hubnum}
\end{figure}

\boldsubsubsection{\texttt{hub\_num}}
We investigate that with large \texttt{hub\_num}, UMATO produces projections with a reliable global structure. Meanwhile, UMATO projections with small \texttt{hub\_num} have a distorted structure, where points are randomly mixed (\autoref{fig:hp}, third row). Intuitively, this is because a small number of hub points may not accurately represent the overall skeletal layout of the original HD data.

To thoroughly examine a sufficient number of \texttt{hub\_num}, we conduct an additional experiment that investigates the accuracy of UMATO projections with different \texttt{hub\_num}. We first generate projections of 20 datasets we use in previous experiments (\autoref{tab:datasets}) while setting \texttt{hub\_num} from 20 to 400 with an interval of 20. We set \texttt{n\_neighbors} to 75 and \texttt{min\_dist} to 0.1, which are the default values of our implementation. We then assess the accuracy of projections using a local metric (F1 score of T\&C with $k=10$) and a global metric (KL divergence with $\sigma=0.1$). Since KL divergence scores closer to 0 indicate more accurate projections, we subtract each KL divergence score from 1 to derive the corresponding value. Finally, to account for varying dataset difficulty, we normalized the scores by dividing each score by the maximum score achieved within its respective dataset.

The result is depicted in \autoref{fig:hubnum}. While the T\&C score converges to its maximum achievable value when \texttt{hub\_num} reaches 200, the KL divergence converges around 350 with greater variance. The results suggest that UMATO demonstrates its maximum ability to preserve local structure with a smaller \texttt{hub\_num} compared to what is needed for preserving global structure. Based on the result, we recommend using \texttt{hub\_num} greater than 200 for analytic tasks focused on local structures (e.g., neighborhood identification) and values greater than 350 for tasks focused on global structures (e.g., cluster density estimation).

%% file: sections/09_discussions.tex
\section{Discussion}

\subsection{Tradeoffs in UMATO}

\label{sec:discuss_tradeoff}

We discuss two prevalent tradeoffs of DR revealed by our study: (1) the tradeoff between local accuracy and global accuracy, and (2) the tradeoff between overall accuracy and running time.

\boldsubsubsection{Tradeoff between local and global accuracy}
UMATO's two-phase optimization scheme brings a clear tradeoff between accuracy in preserving local and global structures.
The scheme aids the algorithm in achieving substantial enhancement in terms of global structure preservation and stability. 
However, the scheme also leads to lower accuracy in depicting local structures (\autoref{sec:accuexp}). 
% This perspective relatively dismisses local structure compared to previous DR techniques. 
UMATO thus may poorly support users in conducting local tasks, e.g., identifying nearest neighbors of a given point \cite{etemadpour15tvcg}.
Here, designing a new DR technique that can explicitly and clearly control the tradeoff between local and global accuracy will be an interesting avenue to explore, as such a technique will allow users ``tune'' their DR projections to align with their task. 
Still, our demonstrations verify that UMATO's balance between local and global structures leads to better preservation of HD structures overall. 
% On the other hand, looking at HD data from this perspective may better support users in conducting global tasks, such as estimating the similarities between clusters. 
In summary, UMATO's strength lies in its ability to illuminate broader patterns in HD data, providing users with more chances to gain new insights.

\boldsubsubsection{Tradeoff between accuracy and runtime}
Another side effect caused by our optimization design is the addition of \texttt{hub\_num}, a hyperparameter that substantially affects final projections. 
The emergence of a new hyperparameter adds additional complexity while using UMATO in practice. 
To alleviate this problem, we provided a guideline to select a proper \texttt{hub\_num}, which is to set a sufficiently large value (\autoref{sec:hp}).
However, we cannot indefinitely raise \texttt{hub\_num} as it will also increase the runtime (\autoref{sec:complexity}). 
To overcome such a tradeoff, we plan to develop an automatic algorithm that finds a good hyperparameter setting \cite{cao2017automatic} that matches a given dataset. 
We also plan to make UMATO more stable against changes in hyperparameters. 
Conducting a large-scale benchmark of UMATO to find the hyperparameter setting that works well, in general, will also be interesting for future work.

\subsection{Limitations and Future Works}

\label{sec:limitations}

We discuss the limitations of this research and possible future works.

\noindent
\textbf{Making UMATO scalable and interactive.}
We believe that UMATO has plenty of room to be improved.
First, UMATO's scalability can be revisited. Currently, UMATO only runs on a CPU, where the main bottleneck is $k$NN computation and local optimization (\autoref{sec:inidividualruntime}). Although our implementation utilizes parallelization based on multithreading, these two stages may be further accelerated using heterogeneous systems, such as GPU \cite{pezzotti2019gpgpu, nolet2020bringing} or FPGA \cite{fernandez2019fpga}.
We can also make the algorithm progressive \cite{ko2020progressive} \rev{or parametric \cite{sainburg21nc} (i.e., be able to project unseen data based on previously trained data)}, making UMATO suitable for responsive visual analytics systems. 
Furthermore, identifying the optimal number of iterations in local optimization will substantially reduce the runtime.
These efforts will help us to add interactivity to UMATO. 
For example, if local optimization can be performed in real-time, we can allow users to steer hub points based on their background knowledge \cite{joia2011local}. 
Second, we do not know whether the current way of selecting hub points ($k$NN-based hub selection; \autoref{sec:ptclassi}) is the optimal way to do so. Investigating alternative ways (e.g., stratified sampling using clustering algorithms \cite{abbas19clustme}) will be an interesting future work.

\noindent
\textbf{Conducting further evaluations.}
We want to evaluate UMATO in detail. For example, we have not yet investigated UMATO's effectiveness in real-world settings. Exploring UMATO's potential in practical applications through a user study will be an interesting future avenue. 
We also plan to conduct a user study evaluating UMATO based on participants' task accuracy \cite{etemadpour15tvcg, xia2021revisiting} or analytical preferences \cite{morariu23tvcg, doh25arxiv}.

\noindent
\textbf{Applying two-phase optimization scheme to other algorithms.}
We verify that the two-phase optimization can improve the global accuracy of DR techniques. Intuitively, we can further investigate whether such a scheme can aid other data abstraction algorithms (e.g., clustering). For example, we may apply UMATO to produce graph layouts (e.g., force-directed layout \cite{hu2005efficient}), as Kruiger et al. \cite{kruiger17cgf} did with $t$-SNE. 

%% file: sections/10_conclusion.tex
\section{Conclusion}

We design and implement a novel DR technique called UMATO.
UMATO divides the optimization of UMAP into two phases, preserving both global and local structures of HD data simultaneously. 
UMATO thereby provides a more faithful visual representation of how manifolds are arranged in the original HD space. 

Our quantitative experiments with diverse real-world datasets validate the accuracy of UMATO in accurately preserving local and global structures (e.g., UMAP and its variants). 
We also qualitatively demonstrate UMATO's accuracy using synthetic datasets.
By providing guidelines for setting hyperparameters and releasing an open-source library, we pave the way for using UMATO in practice.
In summary, our research contributes a significant advancement in the DR research community, opening up opportunities for more reliable and efficient visual analytics.

%% file: main.bbl
\begin{thebibliography}{10}
\providecommand{\url}[1]{#1}
\csname url@samestyle\endcsname
\providecommand{\newblock}{\relax}
\providecommand{\bibinfo}[2]{#2}
\providecommand{\BIBentrySTDinterwordspacing}{\spaceskip=0pt\relax}
\providecommand{\BIBentryALTinterwordstretchfactor}{4}
\providecommand{\BIBentryALTinterwordspacing}{\spaceskip=\fontdimen2\font plus
\BIBentryALTinterwordstretchfactor\fontdimen3\font minus \fontdimen4\font\relax}
\providecommand{\BIBforeignlanguage}[2]{{%
\expandafter\ifx\csname l@#1\endcsname\relax
\typeout{** WARNING: IEEEtran.bst: No hyphenation pattern has been}%
\typeout{** loaded for the language `#1'. Using the pattern for}%
\typeout{** the default language instead.}%
\else
\language=\csname l@#1\endcsname
\fi
#2}}
\providecommand{\BIBdecl}{\relax}
\BIBdecl

\bibitem{jo2018panene}
J.~Jo, J.~Seo, and J.-D. Fekete, ``Panene: A progressive algorithm for indexing and querying approximate k-nearest neighbors,'' \emph{IEEE Transactions on Visualization and Computer Graphics}, vol.~26, no.~2, pp. 1347--1360, 2018.

\bibitem{fujiwara2019supporting}
T.~Fujiwara, O.-H. Kwon, and K.-L. Ma, ``Supporting analysis of dimensionality reduction results with contrastive learning,'' \emph{IEEE Transactions on Visualization and Computer Graphics}, vol.~26, no.~1, pp. 45--55, 2019.

\bibitem{chatzimparmpas2020t}
A.~Chatzimparmpas, R.~M. Martins, and A.~Kerren, ``t-visne: Interactive assessment and interpretation of t-sne projections,'' \emph{IEEE Transactions on Visualization and Computer Graphics}, vol.~26, no.~8, pp. 2696--2714, 2020.

\bibitem{becht2019dimensionality}
E.~Becht, L.~McInnes, J.~Healy, C.-A. Dutertre, I.~W. Kwok, L.~G. Ng, F.~Ginhoux, and E.~W. Newell, ``Dimensionality reduction for visualizing single-cell data using umap,'' \emph{Nature biotechnology}, vol.~37, no.~1, pp. 38--44, 2019.

\bibitem{boggust22iui}
A.~Boggust, B.~Carter, and A.~Satyanarayan, ``Embedding comparator: Visualizing differences in global structure and local neighborhoods via small multiples,'' in \emph{Proceedings of the 27th International Conference on Intelligent User Interfaces}, ser. IUI '22.\hskip 1em plus 0.5em minus 0.4em\relax New York, NY, USA: Association for Computing Machinery, 2022, p. 746–766.

\bibitem{nonato2018multidimensional}
L.~G. Nonato and M.~Aupetit, ``Multidimensional projection for visual analytics: Linking techniques with distortions, tasks, and layout enrichment,'' \emph{IEEE Transactions on Visualization and Computer Graphics}, vol.~25, no.~8, pp. 2650--2673, 2018.

\bibitem{xia2021revisiting}
J.~Xia, Y.~Zhang, J.~Song, Y.~Chen, Y.~Wang, and S.~Liu, ``Revisiting dimensionality reduction techniques for visual cluster analysis: An empirical study,'' \emph{IEEE Transactions on Visualization and Computer Graphics}, vol.~28, no.~1, pp. 529--539, 2021.

\bibitem{silva2003global}
V.~D. Silva and J.~B. Tenenbaum, ``Global versus local methods in nonlinear dimensionality reduction,'' in \emph{Advances in Neural Information Processing Systems}, 2003, pp. 721--728.

\bibitem{mcinnes2018umap}
L.~McInnes, J.~Healy, and J.~Melville, ``Umap: Uniform manifold approximation and projection for dimension reduction,'' \emph{arXiv preprint arXiv:1802.03426}, 2018.

\bibitem{maaten2008visualizing}
L.~v.~d. Maaten and G.~Hinton, ``Visualizing data using t-sne,'' \emph{Journal of machine learning research}, vol.~9, no. Nov, pp. 2579--2605, 2008.

\bibitem{pearson1901liii}
K.~Pearson, ``Liii. on lines and planes of closest fit to systems of points in space,'' \emph{The London, Edinburgh, and Dublin philosophical magazine and journal of science}, vol.~2, no.~11, pp. 559--572, 1901.

\bibitem{tenenbaum2000global}
J.~B. Tenenbaum, V.~De~Silva, and J.~C. Langford, ``A global geometric framework for nonlinear dimensionality reduction,'' \emph{Science}, vol. 290, no. 5500, pp. 2319--2323, 2000.

\bibitem{kruskal64psycho}
J.~Kruskal, ``Multidimensional scaling by optimizing goodness of fit to a nonmetric hypothesis,'' \emph{Psychometrika}, vol.~29, pp. 1--27, 1964.

\bibitem{de2004sparse}
V.~De~Silva and J.~B. Tenenbaum, ``Sparse multidimensional scaling using landmark points,'' technical report, Stanford University, Tech. Rep., 2004.

\bibitem{jeon24tvcg}
H.~Jeon, Y.-H. Kuo, M.~Aupetit, K.-L. Ma, and J.~Seo, ``Classes are not clusters: Improving label-based evaluation of dimensionality reduction,'' \emph{IEEE Transactions on Visualization and Computer Graphics}, vol.~30, no.~1, pp. 781--791, 2024.

\bibitem{jeon2022measuring}
H.~Jeon, H.-K. Ko, J.~Jo, Y.~Kim, and J.~Seo, ``Measuring and explaining the inter-cluster reliability of multidimensional projections,'' \emph{IEEE Transactions on Visualization and Computer Graphics}, vol.~28, no.~1, pp. 551--561, 2022.

\bibitem{fujiwara23pvis}
T.~Fujiwara, Y.-H. Kuo, A.~Ynnerman, and K.-L. Ma, ``Feature learning for nonlinear dimensionality reduction toward maximal extraction of hidden patterns,'' in \emph{2023 IEEE 16th Pacific Visualization Symposium (PacificVis)}, 2023, pp. 122--131.

\bibitem{cashman25tvcg}
D.~Cashman, M.~Keller, H.~Jeon, B.~C. Kwon, and Q.~Wang, ``A critical analysis of the usage of dimensionality reduction in four domains,'' \emph{IEEE Transactions on Visualization and Computer Graphics}, pp. 1--20, 2025.

\bibitem{shneiderman1996eyes}
B.~Shneiderman, ``The eyes have it: A task by data type taxonomy for information visualizations,'' in \emph{Proceedings 1996 IEEE symposium on visual languages}.\hskip 1em plus 0.5em minus 0.4em\relax IEEE, 1996, pp. 336--343.

\bibitem{wang21jmlr}
Y.~Wang, H.~Huang, C.~Rudin, and Y.~Shaposhnik, ``Understanding how dimension reduction tools work: An empirical approach to deciphering t-sne, umap, trimap, and pacmap for data visualization,'' \emph{Journal of Machine Learning Research}, vol.~22, no. 201, pp. 1--73, 2021.

\bibitem{amid2019trimap}
E.~Amid and M.~K. Warmuth, ``Trimap: Large-scale dimensionality reduction using triplets,'' \emph{arXiv preprint arXiv:1910.00204}, 2019.

\bibitem{jeon22vis}
H.~Jeon, H.-K. Ko, S.~Lee, J.~Jo, and J.~Seo, ``Uniform manifold approximation with two-phase optimization,'' in \emph{2022 IEEE Visualization and Visual Analytics (VIS)}, 2022, pp. 80--84.

\bibitem{jeon22visappendix}
------, ``Appendix: Uniform manifold approximation with two-phase optimization,'' 2022.

\bibitem{belkin2002laplacian}
M.~Belkin and P.~Niyogi, ``Laplacian eigenmaps and spectral techniques for embedding and clustering,'' in \emph{Advances in Neural Information Processing Systems}, 2002, pp. 585--591.

\bibitem{mikolov2013distributed}
T.~Mikolov, I.~Sutskever, K.~Chen, G.~S. Corrado, and J.~Dean, ``Distributed representations of words and phrases and their compositionality,'' in \emph{Advances in Neural Information Processing Systems}, 2013, pp. 3111--3119.

\bibitem{tang2015line}
J.~Tang, M.~Qu, M.~Wang, M.~Zhang, J.~Yan, and Q.~Mei, ``Line: Large-scale information network embedding,'' in \emph{Proceedings of the 24th International Conference on World Wide Web}, 2015, pp. 1067--1077.

\bibitem{tang2016visualizing}
J.~Tang, J.~Liu, M.~Zhang, and Q.~Mei, ``Visualizing large-scale and high-dimensional data,'' in \emph{Proceedings of the 25th International Conference on World Wide Web}, 2016, pp. 287--297.

\bibitem{kobak2019umap}
D.~Kobak and G.~C. Linderman, ``Umap does not preserve global structure any better than t-sne when using the same initialization,'' \emph{BioRxiv}, 2019.

\bibitem{lespinats11cgf}
S.~Lespinats and M.~Aupetit, ``Checkviz: Sanity check and topological clues for linear and non-linear mappings,'' \emph{Computer Graphics Forum}, vol.~30, no.~1, pp. 113--125, 2011.

\bibitem{jeon2022distortion}
H.~Jeon, M.~Aupetit, S.~Lee, H.-K. Ko, Y.~Kim, and J.~Seo, ``Distortion-aware brushing for interactive cluster analysis in multidimensional projections,'' \emph{arXiv preprint arXiv:2201.06379}, 2022.

\bibitem{venna2001neighborhood}
J.~Venna and S.~Kaski, ``Neighborhood preservation in nonlinear projection methods: An experimental study,'' in \emph{International Conference on Artificial Neural Networks}.\hskip 1em plus 0.5em minus 0.4em\relax Springer, 2001, pp. 485--491.

\bibitem{lee2007nonlinear}
J.~A. Lee and M.~Verleysen, \emph{Nonlinear dimensionality reduction}.\hskip 1em plus 0.5em minus 0.4em\relax Springer Science \& Business Media, 2007.

\bibitem{hinton2002stochastic}
G.~E. Hinton and S.~Roweis, ``Stochastic neighbor embedding,'' \emph{Advances in Neural Information Processing Systems}, vol.~15, pp. 857--864, 2002.

\bibitem{atzberger24tvcg}
D.~Atzberger, T.~Cech, M.~Trapp, R.~Richter, W.~Scheibel, J.~Döllner, and T.~Schreck, ``Large-scale evaluation of topic models and dimensionality reduction methods for 2d text spatialization,'' \emph{IEEE Transactions on Visualization and Computer Graphics}, vol.~30, no.~1, pp. 902--912, 2024.

\bibitem{espadoto2019towards}
M.~Espadoto, R.~M. Martins, A.~Kerren, N.~S. Hirata, and A.~C. Telea, ``Towards a quantitative survey of dimension reduction techniques,'' \emph{IEEE Transactions on Visualization and Computer Graphics}, 2019.

\bibitem{Moor19Topological}
M.~Moor, M.~Horn, B.~Rieck, and K.~Borgwardt, ``Topological autoencoders,'' in \emph{Proceedings of the 37th International Conference on Machine Learning~(ICML)}, ser. Proceedings of Machine Learning Research.\hskip 1em plus 0.5em minus 0.4em\relax PMLR, 2020.

\bibitem{jackle17vast}
D.~Jäckle, M.~Hund, M.~Behrisch, D.~A. Keim, and T.~Schreck, ``Pattern trails: Visual analysis of pattern transitions in subspaces,'' in \emph{2017 IEEE Conference on Visual Analytics Science and Technology (VAST)}, 2017, pp. 1--12.

\bibitem{martins14cg}
R.~M. Martins, D.~B. Coimbra, R.~Minghim, and A.~Telea, ``Visual analysis of dimensionality reduction quality for parameterized projections,'' \emph{Computers \& Graphics}, vol.~41, pp. 26--42, 2014.

\bibitem{aupetit07neurocomputing}
M.~Aupetit, ``Visualizing distortions and recovering topology in continuous projection techniques,'' \emph{Neurocomputing}, vol.~70, no.~7, pp. 1304--1330, 2007, advances in Computational Intelligence and Learning.

\bibitem{bai2021uibert}
C.~Bai, X.~Zang, Y.~Xu, S.~Sunkara, A.~Rastogi, J.~Chen, and B.~A. y~Arcas, ``Uibert: Learning generic multimodal representations for ui understanding,'' 2021.

\bibitem{lee22arxiv}
J.~W. Lee, E.~Kim, J.~Koo, and K.~Lee, ``Representation selective self-distillation and wav2vec 2.0 feature exploration for spoof-aware speaker verification,'' \emph{arXiv preprint arXiv:2204.02639}, 2022.

\bibitem{lim23chi}
C.~Lim and G.~Park, ``Can a computer tell differences between vibrations?: Physiology-based computational model for perceptual dissimilarity prediction,'' in \emph{Proceedings of the 2023 CHI Conference on Human Factors in Computing Systems}, ser. CHI '23.\hskip 1em plus 0.5em minus 0.4em\relax New York, NY, USA: Association for Computing Machinery, 2023.

\bibitem{narechania22tvcg}
A.~Narechania, A.~Karduni, R.~Wesslen, and E.~Wall, ``Vitality: Promoting serendipitous discovery of academic literature with transformers \& visual analytics,'' \emph{IEEE Transactions on Visualization and Computer Graphics}, vol.~28, no.~1, pp. 486--496, 2022.

\bibitem{hong22pvis}
H.~Hong, S.~Yoo, Y.~Jin, C.~Yoon, S.~Yim, S.~Choi, and Y.~Jang, ``Visual analytics system of comprehensive data quality improvement for machine learning using data- and process-driven strategies,'' in \emph{2022 IEEE International Conference on Big Data (Big Data)}, 2022, pp. 396--401.

\bibitem{kahng18tvcg}
M.~Kahng, P.~Y. Andrews, A.~Kalro, and D.~H. Chau, ``Activis: Visual exploration of industry-scale deep neural network models,'' \emph{IEEE Transactions on Visualization and Computer Graphics}, vol.~24, no.~1, pp. 88--97, 2018.

\bibitem{hinton2006reducing}
G.~E. Hinton and R.~R. Salakhutdinov, ``Reducing the dimensionality of data with neural networks,'' \emph{Science}, vol. 313, no. 5786, pp. 504--507, 2006.

\bibitem{joia2011local}
P.~Joia, D.~Coimbra, J.~A. Cuminato, F.~V. Paulovich, and L.~G. Nonato, ``Local affine multidimensional projection,'' \emph{IEEE Transactions on Visualization and Computer Graphics}, vol.~17, no.~12, pp. 2563--2571, 2011.

\bibitem{yeh09esa}
I.-C. Yeh, K.-J. Yang, and T.-M. Ting, ``Knowledge discovery on rfm model using bernoulli sequence,'' \emph{Expert Systems with Applications}, vol.~36, no.~3, pp. 5866--5871, 2009.

\bibitem{hon2017deep}
M.~Hon, D.~Stello, and J.~Yu, ``Deep learning classification in asteroseismology,'' \emph{Monthly Notices of the Royal Astronomical Society}, vol. 469, no.~4, pp. 4578--4583, 2017.

\bibitem{asuncion07uci}
A.~Asuncion and D.~Newman, ``Uci machine learning repository,'' 2007.

\bibitem{nene20columbia}
S.~Nene, S.~Nayar, H.~Murase \emph{et~al.}, ``Columbia object image library (coil-20), 1996.''

\bibitem{andrzejak2001indications}
R.~G. Andrzejak, K.~Lehnertz, F.~Mormann, C.~Rieke, P.~David, and C.~E. Elger, ``Indications of nonlinear deterministic and finite-dimensional structures in time series of brain electrical activity: Dependence on recording region and brain state,'' \emph{Physical Review E}, vol.~64, no.~6, p. 061907, 2001.

\bibitem{sharan2009material}
L.~Sharan, R.~Rosenholtz, and E.~Adelson, ``Material perception: What can you see in a brief glance?'' \emph{Journal of Vision}, vol.~9, no.~8, pp. 784--784, 2009.

\bibitem{davidson2017automated}
T.~Davidson, D.~Warmsley, M.~Macy, and I.~Weber, ``Automated hate speech detection and the problem of offensive language,'' in \emph{Proceedings of the International AAAI Conference on Web and Social Media}, vol.~11, no.~1, 2017, pp. 512--515.

\bibitem{maas2011learning}
A.~Maas, R.~E. Daly, P.~T. Pham, D.~Huang, A.~Y. Ng, and C.~Potts, ``Learning word vectors for sentiment analysis,'' in \emph{Proceedings of the 49th annual meeting of the association for computational linguistics: Human language technologies}, 2011, pp. 142--150.

\bibitem{kaggle}
``Kaggle,'' \url{https://www.kaggle.com}.

\bibitem{ccinar2020classification}
{\.I}.~{\c{C}}INAR, M.~KOKLU, and {\c{S}}.~TA{\c{S}}DEM{\.I}R, ``Classification of raisin grains using machine vision and artificial intelligence methods,'' \emph{Gazi M{\"u}hendislik Bilimleri Dergisi (GMBD)}, vol.~6, no.~3, pp. 200--209, 2020.

\bibitem{sikora2010application}
M.~Sikora \emph{et~al.}, ``Application of rule induction algorithms for analysis of data collected by seismic hazard monitoring systems in coal mines,'' \emph{Archives of Mining Sciences}, vol.~55, no.~1, pp. 91--114, 2010.

\bibitem{kotzias2015group}
D.~Kotzias, M.~Denil, N.~De~Freitas, and P.~Smyth, ``From group to individual labels using deep features,'' in \emph{Proceedings of the 21th ACM SIGKDD international conference on knowledge discovery and data mining}, 2015, pp. 597--606.

\bibitem{almeida2011contributions}
T.~A. Almeida, J.~M.~G. Hidalgo, and A.~Yamakami, ``Contributions to the study of sms spam filtering: new collection and results,'' in \emph{Proceedings of the 11th ACM symposium on Document engineering}, 2011, pp. 259--262.

\bibitem{ventocilla2020comparative}
E.~Ventocilla and M.~Riveiro, ``A comparative user study of visualization techniques for cluster analysis of multidimensional data sets,'' \emph{Information visualization}, vol.~19, no.~4, pp. 318--338, 2020.

\bibitem{abdelhamid2014phishing}
N.~Abdelhamid, A.~Ayesh, and F.~Thabtah, ``Phishing detection based associative classification data mining,'' \emph{Expert Systems with Applications}, vol.~41, no.~13, pp. 5948--5959, 2014.

\bibitem{zhao25tvcg}
X.~Zhao, S.~Fu, R.~Yang, L.~Yang, Y.~Chen, J.~Zhang, J.~Long, F.~Zhou, and Y.~Zhao, ``Investigating visual perception of degree centrality in graph visualization,'' \emph{IEEE Transactions on Visualization and Computer Graphics}, vol.~31, no.~6, pp. 3679--3692, 2025.

\bibitem{hou20pr}
\BIBentryALTinterwordspacing
J.~Hou, A.~Zhang, and N.~Qi, ``Density peak clustering based on relative density relationship,'' \emph{Pattern Recognition}, vol. 108, p. 107554, 2020. [Online]. Available: \url{https://www.sciencedirect.com/science/article/pii/S0031320320303575}
\BIBentrySTDinterwordspacing

\bibitem{rodriguez14science}
\BIBentryALTinterwordspacing
A.~Rodriguez and A.~Laio, ``Clustering by fast search and find of density peaks,'' \emph{Science}, vol. 344, no. 6191, pp. 1492--1496, 2014. [Online]. Available: \url{https://www.science.org/doi/abs/10.1126/science.1242072}
\BIBentrySTDinterwordspacing

\bibitem{kobak2021initialization}
D.~Kobak and G.~C. Linderman, ``Initialization is critical for preserving global data structure in both t-sne and umap,'' \emph{Nature Biotechnology}, vol.~39, no.~2, pp. 156--157, 2021.

\bibitem{bellman1966dynamic}
R.~Bellman, ``Dynamic programming,'' \emph{Science}, vol. 153, no. 3731, pp. 34--37, 1966.

\bibitem{dong11www}
W.~Dong, C.~Moses, and K.~Li, ``Efficient k-nearest neighbor graph construction for generic similarity measures,'' in \emph{Proceedings of the 20th International Conference on World Wide Web}, ser. WWW '11.\hskip 1em plus 0.5em minus 0.4em\relax New York, NY, USA: Association for Computing Machinery, 2011, p. 577–586.

\bibitem{smelser24beliv}
K.~Smelser, J.~Miller, and S.~Kobourov, ``“normalized stress” is not normalized: How to interpret stress correctly,'' in \emph{2024 IEEE Evaluation and Beyond - Methodological Approaches for Visualization (BELIV)}, 2024, pp. 41--50.

\bibitem{roweis00science}
S.~T. Roweis and L.~K. Saul, ``Nonlinear dimensionality reduction by locally linear embedding,'' \emph{Science}, vol. 290, no. 5500, pp. 2323--2326, 2000.

\bibitem{pedregosa11jmlr}
F.~Pedregosa, G.~Varoquaux, A.~Gramfort, V.~Michel, B.~Thirion, O.~Grisel, M.~Blondel, P.~Prettenhofer, R.~Weiss, V.~Dubourg \emph{et~al.}, ``Scikit-learn: Machine learning in python,'' \emph{the Journal of machine Learning research}, vol.~12, pp. 2825--2830, 2011.

\bibitem{motta23github}
D.~Motta, ``Lmds,'' \url{https://github.com/danilomotta/LMDS}.

\bibitem{jeon23vis}
H.~Jeon, A.~Cho, J.~Jang, S.~Lee, J.~Hyun, H.-K. Ko, J.~Jo, and J.~Seo, ``Zadu: A python library for evaluating the reliability of dimensionality reduction embeddings,'' in \emph{2023 IEEE Visualization and Visual Analytics (VIS)}, 2023, to appear.

\bibitem{chazal2011geometric}
F.~Chazal, D.~Cohen-Steiner, and Q.~M{\'e}rigot, ``Geometric inference for probability measures,'' \emph{Foundations of Computational Mathematics}, vol.~11, no.~6, pp. 733--751, 2011.

\bibitem{snoek12nips}
J.~Snoek, H.~Larochelle, and R.~Adams, ``Practical bayesian optimization of machine learning algorithms,'' in \emph{Advances in Neural Information Processing Systems}, F.~Pereira, C.~Burges, L.~Bottou, and K.~Weinberger, Eds., vol.~25.\hskip 1em plus 0.5em minus 0.4em\relax Curran Associates, Inc., 2012.

\bibitem{lee2009quality}
J.~A. Lee and M.~Verleysen, ``Quality assessment of dimensionality reduction: Rank-based criteria,'' \emph{Neurocomputing}, vol.~72, no. 7-9, pp. 1431--1443, 2009.

\bibitem{venna2010information}
J.~Venna, J.~Peltonen, K.~Nybo, H.~Aidos, and S.~Kaski, ``Information retrieval perspective to nonlinear dimensionality reduction for data visualization.'' \emph{Journal of Machine Learning Research}, vol.~11, no.~2, 2010.

\bibitem{Coenen2019}
A.~Coenen and A.~Pearce, ``Understanding umap,'' \url{https://pair-code.github.io/understanding-umap/}, 2019.

\bibitem{lewis2004rcv1}
D.~D. Lewis, Y.~Yang, T.~Russell-Rose, and F.~Li, ``Rcv1: A new benchmark collection for text categorization research,'' \emph{Journal of machine learning research}, vol.~5, no. Apr, pp. 361--397, 2004.

\bibitem{boukerche20cs}
\BIBentryALTinterwordspacing
A.~Boukerche, L.~Zheng, and O.~Alfandi, ``Outlier detection: Methods, models, and classification,'' \emph{ACM Comput. Surv.}, vol.~53, no.~3, Jun. 2020. [Online]. Available: \url{https://doi.org/10.1145/3381028}
\BIBentrySTDinterwordspacing

\bibitem{jung2025arxiv}
\BIBentryALTinterwordspacing
M.~Jung, T.~Fujiwara, and J.~Jo, ``Ghostumap2: Measuring and analyzing (r,d)-stability of umap,'' 2025. [Online]. Available: \url{https://arxiv.org/abs/2507.17174}
\BIBentrySTDinterwordspacing

\bibitem{etemadpour15tvcg}
R.~Etemadpour, R.~Motta, J.~G. d.~S. Paiva, R.~Minghim, M.~C.~F. de~Oliveira, and L.~Linsen, ``Perception-based evaluation of projection methods for multidimensional data visualization,'' \emph{IEEE Transactions on Visualization and Computer Graphics}, vol.~21, no.~1, pp. 81--94, 2015.

\bibitem{lu20tvcg}
M.~Lu, S.~Wang, J.~Lanir, N.~Fish, Y.~Yue, D.~Cohen-Or, and H.~Huang, ``Winglets: Visualizing association with uncertainty in multi-class scatterplots,'' \emph{IEEE Transactions on Visualization and Computer Graphics}, vol.~26, no.~1, pp. 770--779, 2020.

\bibitem{yunhai18tvcg}
Y.~Wang, K.~Feng, X.~Chu, J.~Zhang, C.-W. Fu, M.~Sedlmair, X.~Yu, and B.~Chen, ``A perception-driven approach to supervised dimensionality reduction for visualization,'' \emph{IEEE Transactions on Visualization and Computer Graphics}, vol.~24, no.~5, pp. 1828--1840, 2018.

\bibitem{xiao2017-online}
H.~Xiao, K.~Rasul, and R.~Vollgraf. (2017) Fashion-mnist: a novel image dataset for benchmarking machine learning algorithms.

\bibitem{wattenberg2016how}
\BIBentryALTinterwordspacing
M.~Wattenberg, F.~Viégas, and I.~Johnson, ``How to use t-sne effectively,'' \emph{Distill}, 2016. [Online]. Available: \url{http://distill.pub/2016/misread-tsne}
\BIBentrySTDinterwordspacing

\bibitem{cao2017automatic}
Y.~Cao and L.~Wang, ``Automatic selection of t-sne perplexity,'' \emph{arXiv preprint arXiv:1708.03229}, 2017.

\bibitem{pezzotti2019gpgpu}
N.~Pezzotti, J.~Thijssen, A.~Mordvintsev, T.~H{\"o}llt, B.~Van~Lew, B.~P. Lelieveldt, E.~Eisemann, and A.~Vilanova, ``Gpgpu linear complexity t-sne optimization,'' \emph{IEEE Transactions on Visualization and Computer Graphics}, vol.~26, no.~1, pp. 1172--1181, 2019.

\bibitem{nolet2020bringing}
C.~J. Nolet, V.~Lafargue, E.~Raff, T.~Nanditale, T.~Oates, J.~Zedlewski, and J.~Patterson, ``Bringing umap closer to the speed of light with gpu acceleration,'' \emph{arXiv preprint arXiv:2008.00325}, 2020.

\bibitem{fernandez2019fpga}
D.~Fernandez, C.~Gonzalez, D.~Mozos, and S.~Lopez, ``Fpga implementation of the principal component analysis algorithm for dimensionality reduction of hyperspectral images,'' \emph{Journal of Real-Time Image Processing}, vol.~16, pp. 1395--1406, 2019.

\bibitem{ko2020progressive}
H.-K. Ko, J.~Jo, and J.~Seo, ``Progressive uniform manifold approximation and projection,'' in \emph{22nd Eurographics Conference on Visualization, EuroVis 2020-Short Papers}.\hskip 1em plus 0.5em minus 0.4em\relax Eurographics Association, 2020, pp. 133--137.

\bibitem{sainburg21nc}
T.~Sainburg, L.~McInnes, and T.~Q. Gentner, ``{Parametric UMAP Embeddings for Representation and Semisupervised Learning},'' \emph{Neural Computation}, vol.~33, no.~11, pp. 2881--2907, 10 2021.

\bibitem{abbas19clustme}
M.~M. Abbas, M.~Aupetit, M.~Sedlmair, and H.~Bensmail, ``Clustme: A visual quality measure for ranking monochrome scatterplots based on cluster patterns,'' \emph{Computer Graphics Forum}, vol.~38, no.~3, pp. 225--236, 2019.

\bibitem{morariu23tvcg}
C.~Morariu, A.~Bibal, R.~Cutura, B.~Frénay, and M.~Sedlmair, ``Predicting user preferences of dimensionality reduction embedding quality,'' \emph{IEEE Transactions on Visualization and Computer Graphics}, vol.~29, no.~1, pp. 745--755, 2023.

\bibitem{doh25arxiv}
S.~Doh, H.~Jeon, S.~Shin, G.~J. Quadri, N.~W. Kim, and J.~Seo, ``Understanding bias in perceiving dimensionality reduction projections,'' \emph{arXiv preprint arXiv:2507.20805}, 2025.

\bibitem{hu2005efficient}
Y.~Hu, ``Efficient, high-quality force-directed graph drawing,'' \emph{Mathematica journal}, vol.~10, no.~1, pp. 37--71, 2005.

\bibitem{kruiger17cgf}
J.~F. Kruiger, P.~E. Rauber, R.~M. Martins, A.~Kerren, S.~Kobourov, and A.~C. Telea, ``Graph layouts by t-sne,'' \emph{Computer Graphics Forum}, vol.~36, no.~3, pp. 283--294, 2017.

\end{thebibliography}
